\documentclass[letterpaper, 10 pt, journal, twoside]{ieeetran}
\IEEEoverridecommandlockouts
\usepackage[pdftex]{graphicx}
\graphicspath{{images}}
\DeclareGraphicsExtensions{.pdf,.jpeg,.png}

\usepackage{amsmath}
\usepackage{array}

\usepackage[font=small]{caption}
\usepackage{subcaption}
\usepackage{setspace}
\let\labelindent\relax
\usepackage{enumitem}
\usepackage{url}
\usepackage{amssymb}
\usepackage{bm}
\usepackage{graphicx}
\usepackage{float}
\usepackage[normalem]{ulem}
\usepackage{listings}
\usepackage{gensymb}
\usepackage{array}
\usepackage{pbox}
\usepackage{mathtools}
\usepackage{breqn}
\usepackage[ruled,vlined]{algorithm2e}
\usepackage{resizegather}
\usepackage{changepage}
\usepackage{titlesec}
\usepackage{geometry}
\titlespacing\section{0pt}{8pt plus 2pt minus 2pt}{5pt plus 2pt minus 2pt}
\titlespacing\subsection{0pt}{8pt minus 2pt}{5pt minus 2pt}
\newlength{\offsetpage}
\newenvironment{widepage}{\begin{adjustwidth}{-\offsetpage}{-\offsetpage}%
    \addtolength{\textwidth}{2\offsetpage}}%
{\end{adjustwidth}}
\usepackage{xpatch}
\xpretocmd{\algorithm}{\hsize=\linewidth}{}{}
\graphicspath{
    {images/}
}
\newcommand{\figref}{Fig.~\ref}
\newcommand{\tabref}{Table~\ref}

\newcommand{\mbs}{\bm}
\newcommand{\mbf}{\mathbf}

\newcommand{\mbc}{\mathcal}
\newcommand{\bbm}{\begin{bmatrix}}
\newcommand{\ebm}{\end{bmatrix}}
\newcommand{\col}{\mathbin{:}}
\DeclareMathOperator*{\argmin}{\arg\!\min}

\newcolumntype{L}{>{$}l<{$}}

\usepackage{xcolor}
\makeatletter
\newcommand{\removelatexerror}{\let\@latex@error\@gobble}
\makeatother

\begin{document}

\title{Koopman Linearization for Data-Driven Batch State Estimation of Control-Affine Systems}

\author{Zi Cong Guo$^{1}$, Vassili Korotkine$^{2}$, James R. Forbes$^{2}$, and Timothy D. Barfoot$^{1}$
\thanks{Manuscript received September 9, 2021; Revised November 23, 2021; Accepted November 24, 2021.} %Use only for final RAL version
\thanks{This paper was recommended for publication by Editor S. Behnke upon evaluation of the Associate Editor and Reviewers' comments.
This work was generously supported by the National Sciences and Engineering Research Council (NSERC), the Canadian Institute for Advanced Research (CIFAR), the Canada Foundation for Innovation (CFI), and the Fonds de recherche du Qu\'{e}bec (FRQNT).} %Use only for final RAL version
\thanks{$^{1}$Zi Cong Guo and Timothy D. Barfoot are with the University of Toronto Institute for Aerospace Studies, University of Toronto, Toronto, Ontario, Canada  \tt\footnotesize{zc.guo@mail.utoronto.ca}}
\thanks{$^{2}$Vassili Korotkin and James R. Forbes are with the Department of Mechanical Engineering, McGill University, Montreal, Quebec, Canada  \tt\footnotesize{vassili.korotkine@mail.mcgill.ca}}
% \thanks{Digital Object Identifier (DOI): see top of this page.}
}
% Use only for final RAL version.

\maketitle

\begin{abstract}
We present the Koopman State Estimator (KoopSE), a framework for model-free batch state estimation of control-affine systems that makes no linearization assumptions, requires no problem-specific feature selections, and has an inference computational cost that is independent of the number of training points. We lift the original nonlinear system into a higher-dimensional Reproducing Kernel Hilbert Space (RKHS), where the system becomes bilinear. The time-invariant model matrices can be learned by solving a least-squares problem on training trajectories. At test time, the system is algebraically manipulated into a linear time-varying system, where standard batch linear state estimation techniques can be used to efficiently compute state means and covariances. Random Fourier Features (RFF) are used to combine the computational efficiency of Koopman-based methods and the generality of kernel-embedding methods. KoopSE is validated experimentally on a localization task involving a mobile robot equipped with ultra-wideband receivers and wheel odometry. KoopSE estimates are more accurate and consistent than the standard model-based extended Rauch–Tung–Striebel (RTS) smoother, despite KoopSE having no prior knowledge of the system's motion or measurement models.

% \vspace{10pt}
% {\textbf{\textit{Index Terms}---Localization, Probabilistic Inference.}}
\end{abstract}

\begin{IEEEkeywords}
Localization, Probabilistic Inference.
\end{IEEEkeywords}
\IEEEpeerreviewmaketitle
\section{Introduction}
\IEEEPARstart{S}{tate} estimation is an essential component of almost all robotic systems. Having accurate and consistent state estimates not only assists with high-level decision-making, but also directly improves the performance of other modules, such as planning and control. While there are many established techniques of estimation for linear-Gaussian systems, classical techniques for nonlinear state estimation still pose challenges, requiring complex modelling of the system and/or problem-specific estimation techniques. Model-free estimation algorithms, in contrast, learn key aspects of the system models from training data, then perform estimation on similarly distributed test data. They can be applied to a wider variety of systems without requiring a priori models. One can learn the models directly with neural network-based estimators such as differentiable filters \cite{diff-filter}, but a powerful alternative approach involves lifting the system into a higher-dimensional space, either by embedding probability distributions in a Reproducing Kernel Hilbert Space (RKHS) or by approximating the Koopman operator. However, kernel embedding methods suffer from poor scalability, and Koopman-based methods usually require problem-specific basis functions. 
Furthermore, most Koopman-based methods learn a lifted linear realization of the system, but many systems in robotics are control-affine, which has a lifted bilinear realization but not necessarily a linear one \cite{control-affine-to-bilin}. As well, to the best of our knowledge, there are no data-driven algorithms that formulate into a clean batch framework of linear state estimation, thus limiting their uses in robotic applications.
% image of robot
\begin{figure}[!t]
 \centering
 \begin{widepage}
 \hspace{-3pt}
  \subcaptionbox*{}{
\includegraphics[width=0.465\columnwidth,trim={0 0 0 0},clip]{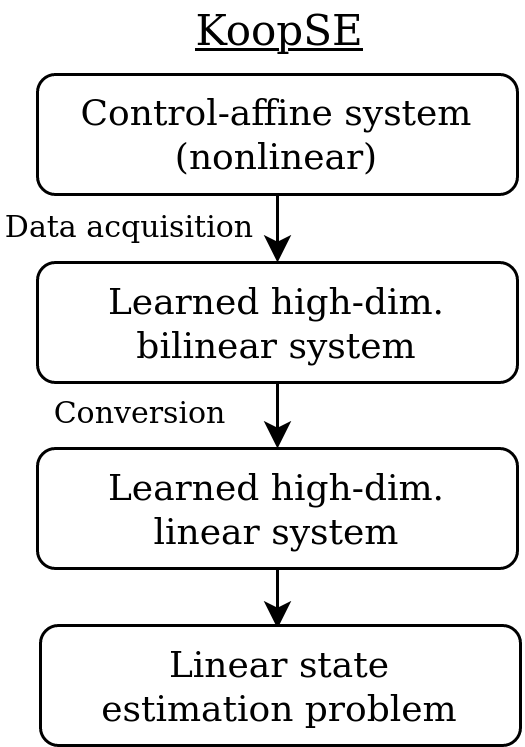}
   }
 \subcaptionbox*{}{
\includegraphics[width=0.48\columnwidth,trim={0 0 0 0},clip]{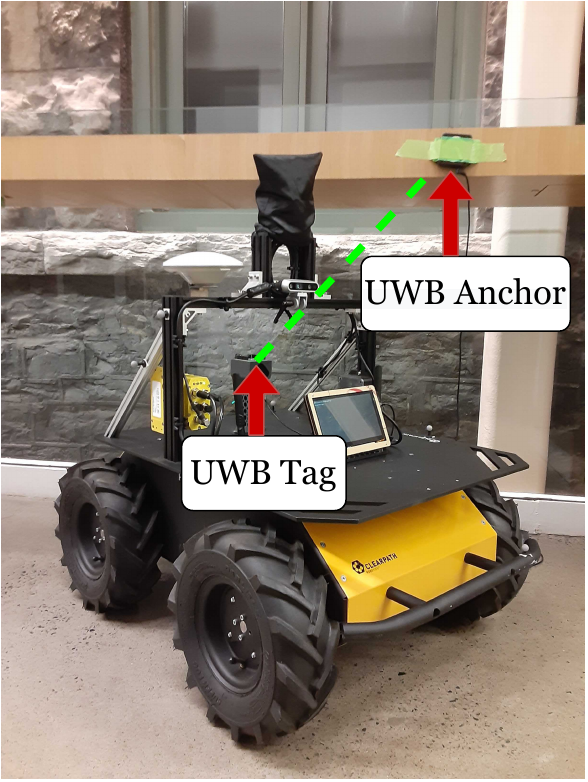}
   }
   \end{widepage}
 \vspace{-10pt}
 \caption{Left: KoopSE concept flowchart. KoopSE uses training data to learn a high-dimensional system, allowing inference to be done by solving a linear state estimation problem. Right: Experimental setup for validating KoopSE. A Husky robot drives around in an indoor environment while receiving range measurements from ultra-wideband (UWB) anchors and logging wheel odometry. }
 \label{fig:robot-pic}
\end{figure}
In this work, we propose the Koopman State Estimator (KoopSE), a state estimation framework that combines aspects from both kernel embedding and Koopman-based methods in a novel way such that it
\begin{itemize}%[leftmargin=13pt]
 \item is applicable to control-affine systems with no prior knowledge of the process and measurement models,
 \item formulates the problem as a high-dimensional batch linear state estimation framework, which admits solutions for state means and covariances, and
 \item has a training cost that scales linearly with the amount of data and an inference cost that is independent of the amount of training data, in contrast with standard kernel methods. 
\end{itemize}

This paper is structured as follows. After reviewing related work in Section \ref{sec:related-work}, we summarize theories for kernel embeddings and Koopman for control-affine systems in Section \ref{sec:prelim}. We derive KoopSE through Sections \ref{sec:system-id}-\ref{sec:recover_estimates}, then apply the RFF approximation in Section \ref{sec:approx-with-rff}. We present experimental results in Sections \ref{sec:experiments} and conclude in Section \ref{sec:discussion_conclusion}. 

\section{Related Work}
\label{sec:related-work}
Methods involving kernels and RKHS are becoming widely used in machine learning and robotics \cite{Hofmann}, allowing for model-free state estimation. Kernel mean embeddings allow probability distributions to be embedded as elements of a RKHS, allowing for operations on random variables in a higher-dimensional space \cite{song}. The kernelized version of Bayes' Rule was first used by \cite{kbr} to build the kernelized Bayes filter, which was subsequently extended to a kernelized smoother by \cite{ksmoother}. Although these methods can exactly learn nonlinear systems, they scale cubically with the number of training samples. To reduce the computational cost, \cite{kkr} used regularization assumptions and subspace projections to construct the Kernel Kalman Filter, which instead scales linearly with training data. Kernel embedding methods in general have prediction cost that scales poorly with training data as well as only being applicable to systems without control inputs. Other kernelized methods such as Gaussian Process (GP) regression \cite{gp-book} have also been used \cite{GPBayes}, but GPs assume additive Gaussian noises on the system models, performing poorly compared to kernel embedding methods for systems with multimodal noise \cite{GPnoiseMulti}.

In contrast to kernel-embedding methods, Koopman-based methods work in the feature space directly, lifting the states into higher dimensions using a set of basis functions (i.e., features) \cite{Koopman}, \cite{mauroy_2020_koopman}. The system matrices can be computed efficiently using extended Dynamic Mode Decomposition (EDMD) \cite{dmd-book}, \cite{dmd-big-book}, and cost of inference is constant with respect to training points at test time. However, most techniques use basis functions tailored to their specific problems \cite{koopman-so3}, \cite{koopman-control}, limiting their generality. It was shown by \cite{control-affine-to-bilin} that a deterministic control-affine system can be exactly represented as a lifted bilinear system using a Koopman operator, and the choice of basis functions can be taken as polynomial, Fourier, or other generic sets of features. Although \cite{control-affine-to-bilin} did not consider sensor measurements nor state covariances, similar to the majority of Koopman-based methods, this equivalence will be foundational for our KoopSE framework.

A variant of Fourier features is Random Fourier Features (RFF) \cite{rff}, a set of probabilistic features for approximating the associated kernel functions. RFF has been effective for various robotics problems, such as continuous occupancy mapping \cite{hilbert-maps} and robot dynamics learning \cite{rff-learn-dynamics}. Although \cite{koopman-with-rff} demonstrated that RFF can feasibly approximate the Koopman operator, the features were again used directly in EDMD. We instead leverage the connection between RFF and kernel embeddings in our work, combining the efficiency of Koopman-based methods with the generality of kernel-embedding methods.

\section{Preliminaries}
\label{sec:prelim}

\subsection{Reproducing Kernel Hilbert Space (RKHS) Embeddings}
\label{sec:rkhs-embeddings}
Consider a general nonlinear system of the form
\pagebreak
\begin{subequations}\label{eq:general-system}
\begin{align}
\mbs{\xi}_k & = \mbf{f}( \mbs{\xi}_{k-1}, \mbs{\nu}_k, \mbs{\omega}_k ), \\
\mbs{\gamma}_k & = \mbf{g}(\mbs{\xi}_k, \mbs{\eta}_k ),
\end{align}
\end{subequations}
where $\mbs{\xi}_k \in \mathbb{R}^{N_\xi}$ is the state, $\mbs{\nu}_k \in \mathbb{R}^{N_\nu}$ the control input, $\mbs{\omega}_k \in \mathbb{R}^{N_\omega}$ the process noise, $\mbs{\gamma}_k \in \mathbb{R}^{N_\gamma}$ the measurement output, and $\mbs{\eta}_k \in \mathbb{R}^{N_\eta}$ the measurement noise, all at timestep $k$. We embed each of the state, input, and measurements in an appropriate RKHS \cite{rkhs}, \cite{rkhs-book},
\begin{gather}
\label{eq:embeddings}
\mbf{x}_k = \mbf{x}( \mbs{\xi}_k ), \quad
\mbf{u}_k  = \mbf{u}(\mbs{\nu}_k), \quad
\mbf{y}_k = \mbf{y}(\mbs{\gamma}_k),
\end{gather}
where $\mbf{x}: \mathbb{R}^{N_\xi}\rightarrow\ \mathcal{X}$, $\mbf{u}: \mathbb{R}^{N_\nu}\rightarrow \mathcal{U}$, $\mbf{y}: \mathbb{R}^{N_\gamma}\rightarrow \mathcal{Y}$ are the possibly infinite-dimensional embeddings (feature maps) associated with the kernels of $\mathcal{X}$, $\mathcal{U}$, and $\mathcal{Y}$, respectively. Unlike Koopman-based methods, which do not restrict the embedding space type, a RKHS embeds entire distributions of random variables in the higher-dimensional space. The distribution is embedded by the {\em mean map} \cite{kernel-embeddings},
\begin{equation}
 \mbs{\mu}_k = \mathbb{E}[\mbf{x}_k],
\end{equation}
with $\mathbb{E}[\cdot]$ as the expectation operator. So long as the kernel associated with $\mbc{X}$ is {\em characteristic} then the mean map can represent any distribution over $\mbs{\xi}_k$.  Given a finite set of samples of random variable, $\mbs{\xi}$, we can approximate the mean map as
\begin{equation}
\mbs{\mu}_k \approx \sum_{i=0}^M m_{k,i} \mbf{x}_i,
\end{equation}
where the weights $m_{k,i}$ depend on how the samples were drawn. We can also rewrite this as $\mbs{\mu}_k \approx \mbf{X} \mbf{m}_k$, where
\begin{equation}
\mbf{X} = \begin{bmatrix} \mbf{x}_{0} & \cdots & \mbf{x}_{M} \end{bmatrix}, \quad \mbf{m}_k = \begin{bmatrix} m_{k,0} & \cdots & m_{k,M} \end{bmatrix}^T.
\end{equation}
We see the samples, $\mbf{X}$, acting as a basis for $\mbc{X}$ with weights $\mbf{m}_k$.  If we then want to calculate the expectation of any function of $\mbs{\xi}$, the mean map allows us to do this as $\mathbb{E}[ h(\mbs{\xi}) ] \approx \sum_{i=0}^M m_{k,i} h(\mbs{\xi}_{i})$ for some nonlinear function $h(\cdot)$.  In particular, if $h(\cdot)$ is the identity function then we have
\begin{equation}
\mathbb{E}[ \mbs{\xi}_k ] \approx \sum_{i=0}^M m_{k,i} \mbs{\xi}_{i} = \mbs{\Xi} \,\mbf{m}_k,
\end{equation}
as expected where $\mbs{\Xi} = \begin{bmatrix} \mbs{\xi}_0 & \cdots & \mbs{\xi}_M \end{bmatrix}$. This is the Representer Theorem \cite{representer-thm}, and we can use this property to recover the mean of our random variable in the original space after doing calculations in the RKHS, eliminating the need to find inverse transformations from the lifted to the original space.

In addition, we can define an {\em centered covariance map} \cite{kernel-embeddings} similarly to the mean map as
\begin{equation}
\mbs{\Sigma}_{k \ell} = \mathbb{E}\left[ (\mbf{x}_k-\mbs{\mu}_k)( \mbf{x}_\ell - \mbs{\mu}_\ell)^T \right],
\end{equation}
which can also represent any joint distribution over $(\mbs{\xi}_k, \mbs{\xi}_\ell)$ provided the kernel associated with $\mbc{X}$ is characteristic.  Given a finite set of samples from  $(\mbs{\xi}_k, \mbs{\xi}_\ell)$, we can approximate the covariance map as $\mbs{\Sigma}_{k\ell} \approx \mbf{X} \mbf{S}_{k\ell} \mbf{X}^T$, where $\mbf{S}_{k\ell}$ is a weight matrix whose values depend on how the samples were drawn. Unlike Koopman, these maps allow for the recovery of covariances of random variables from the lifted space.

The main idea that we will pursue in this paper is to use the embeddings in \eqref{eq:embeddings} to embed a control-affine system in an RKHS where we can write it as a bilinear system. In Sections \ref{sec:system-id} and \ref{sec:batch-est}, we assume that we can explicitly lift quantities into their respective RKHS embeddings, even though $\mbf{x}(\cdot), \mbf{u}(\cdot), \mbf{y}(\cdot)$ are potentially infinite dimensional. We will later see in Section \ref{sec:approx-with-rff} how these derivations still hold under RFF approximations of the feature maps.
\vspace{-5pt}
\subsection{Lifting of Control-Affine Systems}
\label{sec:lifting-control-affine}
Many robotics systems can be written in control-affine form affected by process and measurement noise,
\begin{subequations}\label{eq:control-affine}
\begin{align}
 \mbs{\xi}_{k} &= \mbf{f}_0(\mbs{\xi}_{k-1}) + \sum_{i=1}^{N_\nu} \mbf{f}_i(\mbs{\xi}_{k-1})\nu_{k,i} + \mbs{\omega}_k, \\
 \mbs{\gamma}_k &= \mbf{g}(\mbs{\xi}_k,\mbs{\eta}_k),
\end{align}
\end{subequations}
where $\mbs{\xi}_k\in\mathbb{R}^{N_\xi}$,
$\mbs{\nu}_k = \begin{bmatrix} \nu_{k,1} & \cdots & \nu_{k,N_\nu} \end{bmatrix}^T\in\mathbb{R}^{N_\nu}$,
$\mbs{\omega}_k\in\mathbb{R}^{N_\xi}$,
$\mbs{\gamma}_k\in\mathbb{R}^{N_\gamma}$,
$\mbs{\eta}_k\in\mathbb{R}^{N_\eta}$ represents the same quantities as in the general nonlinear system in \eqref{eq:general-system}, and $\mbf{f}_i$ are the various components of the dynamic model. We look to lift this system into an RKHS. With a sufficiently rich feature map, $\mbf{x}(\cdot)$, a deterministic control-affine motion model (i.e., $\mbs{\omega}_k=\mbf{0}$) can be written exactly as a bilinear model in a lifted space \cite{control-affine-to-bilin},
\begin{equation}\label{eq:bilin-determ}
  \mbf{x}_{k} = \mbf{A}\mbf{x}_{k-1} + \mbf{B}\mbs{\nu}_k + \mbf{H} \left(\mbs{\nu}_k \otimes \mbf{x}_{k-1} \right),
\end{equation}
where $\otimes$ represents the tensor product, equivalent to the Kronecker product if $\mathcal{X}$ is finite-dimensional. We assume that this result holds fairly well for a stochastic system, where the lifted noise becomes additive and Gaussian. This is reasonable as the combination of various sources of random and systematic errors in very high dimensions likely approaches a Gaussian under the Central Limit Theorem. The original equivalence by \cite{control-affine-to-bilin} used the unlifted control input, $\mbs{\nu}_k$, in the lifted space, but we will use its lifted counterpart, $\mbf{u}_k$, since the equivalence still holds if $\mbf{u}(\cdot)$ is sufficiently rich. 

For the measurement model, we assume that the lifted model is linear in the deterministic case,
\begin{equation}
 \mbs{\gamma}_k = \mbf{g}(\mbs{\xi}_k, \mbs{\eta}_k=\mbf{0}) \; \Rightarrow \; \mbf{y}_k = \mbf{C}\mbf{x}_k,
\end{equation}
and we make a similar additive-Gaussian assumption for the measurement noise. The resulting time-invariant stochastic bilinear system in the lifted space can be written as
\begin{subequations}
\label{eq:bilin-stochastic}
\begin{align}
 \mbf{x}_{k} &= \mbf{A}\mbf{x}_{k-1} + \mbf{B}\mbf{u}_k + \mbf{H} \left(\mbf{u}_k \otimes \mbf{x}_{k-1} \right) + \mbf{w}_k, \\
 \mbf{y}_k & = \mbf{C} \mbf{x}_k + \mbf{n}_k,
\end{align}
\end{subequations}
where $\mbf{w}_k \sim \mathcal{N}(\mbf{0},\mbf{Q})$ and $\mbf{n}_k \sim \mathcal{N}(\mbf{0}, \mbf{R})$ are the process and measurement noises, respectively. We also have
 \begin{subequations}
  \begin{gather}
   \mbf{w}_k \in \mathcal{X}, \quad
   \mbf{n}_k \in \mathcal{Y}, \quad
   \mbf{Q} \in \mathcal{X} \times \mathcal{X}, \\
   \mbf{R} \in \mathcal{Y} \times \mathcal{Y}, \quad
   \mbf{A} \col \mathcal{X} \to \mathcal{X}, \quad
   \mbf{B} \col \mathcal{U} \to \mathcal{X}, \\
   \mbf{H} \col \mathcal{U} \otimes \mathcal{X} \to \mathcal{X}, \quad
   \mbf{C} \col \mathcal{X} \to \mathcal{Y}.
  \end{gather}
 \end{subequations}
This lifted system with a bilinear motion model and a linear measurement model is significantly easier to work with than the general control-affine system in \eqref{eq:control-affine}. However, since we assumed the system model is not given in either form (original or lifted), we first need a method of learning the lifted model from data.

\section{System Identification}
\label{sec:system-id}

\subsection{Lifted Matrix Form of Dataset}
Our objective is to learn the lifted system matrices $\mbf{A},\mbf{B},\mbf{H},\mbf{C},\mbf{Q},\mbf{R}$ in \eqref{eq:bilin-stochastic} from data. To this end, we assume a dataset of the control-affine system, including the ground-truth state transitions with their associated control inputs and measurements for $P$ states: $\{ \tilde{\mbs{\xi}}^{(i)}, \mbs{\xi}^{(i)}, \mbs{\nu}^{(i)}, \mbs{\gamma}^{(i)} \}_{i=1}^P$. Here, $\tilde{\mbs{\xi}}^{(i)}$ transitions to $\mbs{\xi}^{(i)}$ under input $\mbs{\nu}^{(i)}$, and receives a measurement $\mbs{\gamma}^{(i)}$ at $\mbs{\xi}^{(i)}$. This format allows for data from one or multiple training trajectories to be used at once. If the dataset consists of a single trajectory of $P+1$ states and $i$ represents the timestep, then we would set $\tilde{\mbs{\xi}}^{(i)} = \mbs{\xi}^{(i-1)}$. In any case, we write the data neatly in block-matrix form:
\begin{subequations}\label{eq:block-matrix-train}
\begin{alignat}{2}
 \mbs{\Xi} &= \begin{bmatrix} \mbs{\xi}^{(1)} & \cdots & \mbs{\xi}^{(P)} \end{bmatrix}, \quad
 &\tilde{\mbs{\Xi}} &= \begin{bmatrix} \tilde{\mbs{\xi}}^{(1)} & \cdots & \tilde{\mbs{\xi}}^{(P)} \end{bmatrix}, \\
 \mbs{\Gamma} &= \begin{bmatrix} \mbs{\gamma}^{(1)} & \cdots & \mbs{\gamma}^{(P)} \end{bmatrix}, \quad
 &\mbs{\Upsilon} &= \begin{bmatrix} \mbs{\nu}^{(1)} & \cdots & \mbs{\nu}^{(P)} \end{bmatrix}.
\end{alignat}
\end{subequations}
The data translates to $\{ \tilde{\mbf{x}}^{(i)}, \mbf{x}^{(i)}, \mbf{u}^{(i)}, \mbf{y}^{(i)} \}_{i=1}^P$ in the lifted space such that
\begin{subequations}
\begin{align}
\mbf{x}^{(i)} &= \mbf{A}\tilde{\mbf{x}}^{(i)} + \mbf{B}\mbf{u}^{(i)} + \mbf{H} \left(\mbf{u}^{(i)} \otimes \tilde{\mbf{x}}^{(i)} \right) + \mbf{w}^{(i)},\\
\mbf{y}^{(i)} &= \mbf{C} \mbf{x}^{(i)} + \mbf{n}^{(i)},
\end{align}
\end{subequations}
for some unknown noise, $\mbf{w}^{(i)}\sim\mathcal{N}(\mbf{0},\mbf{Q})$, $\mbf{n}^{(i)} \sim \mathcal{N}(\mbf{0}, \mbf{R})$. We rewrite the lifted versions of the data and the noises in block-matrix form:
\begin{subequations}
{\small
\begin{align}
 \mbf{X} &= \begin{bmatrix} \mbf{x}^{(1)} & \cdots & \mbf{x}^{(P)} \end{bmatrix}, \; 
 &\tilde{\mbf{X}} = \begin{bmatrix} \tilde{\mbf{x}}^{(1)} & \cdots & \tilde{\mbf{x}}^{(P)} \end{bmatrix}, \\
 \mbf{Y} &= \begin{bmatrix} \mbf{y}^{(1)} & \cdots & \mbf{y}^{(P)} \end{bmatrix}, \;
 &\mbf{U} = \begin{bmatrix} \mbf{u}^{(1)} & \cdots & \mbf{u}^{(P)} \end{bmatrix}, \\
 \mbf{W} &= \begin{bmatrix}  \mbf{w}^{(1)} & \cdots & \mbf{w}^{(P)} \end{bmatrix}, \;
 &\mbf{N} = \begin{bmatrix} \mbf{n}^{(1)} & \cdots & \mbf{n}^{(P)} \end{bmatrix}.
\end{align}
}
\end{subequations}
\\[-10pt]  % weird space after \small if line break is not used
The lifted matrix form of the system for this dataset is
\begin{subequations}
\begin{align}
\mbf{X} &= \mbf{A}\tilde{\mbf{X}} + \mbf{B}\mbf{U} + \mbf{H}\left(\mbf{U} \odot \tilde{\mbf{X}} \right) + \mbf{W}, \\
\mbf{Y} &= \mbf{C}\mbf{X} + \mbf{N},
\end{align}
\end{subequations}
where $\odot$ denotes the Khatri-Rao (column-wise) tensor product.

\subsection{Loss function}
\label{sec:loss-fn}

We now design a loss function from which to optimize for the system matrices from the data. Rather than using the EDMD framework of forming the matrices with major Koopman modes, we use Tikhonov regularization for a cleaner formulation. The model learning problem is posed as
\begin{gather}
 \left\{\mbf{A}^\star,\mbf{B}^\star,\mbf{H}^\star,\mbf{C}^\star,\mbf{Q}^\star,\mbf{R}^\star\right\} = \argmin_{\{\mbf{A},\mbf{B},\mbf{H},\mbf{C},\mbf{Q},\mbf{R}\}} V(\mbf{A},\mbf{B},\mbf{H},\mbf{C},\mbf{Q},\mbf{R}),
\end{gather}
where the loss function, $V=V_1+V_2$, is the sum of 
\begin{subequations}
\begin{align}
% V_1
V_1 =& \frac{1}{2} \left\| \mbf{X} - \mbf{A} \tilde{\mbf{X}} - \mbf{B} \mbf{U} - \mbf{H}\left( \mbf{U} \odot \tilde{\mbf{X}} \right) \right\|^2_{\mbf{Q}^{-1}} \nonumber \\
&+ \frac{1}{2} \left\| \mbf{Y} - \mbf{C} \mbf{X} \right\|^2_{\mbf{R}^{-1}} - \frac{1}{2} {P} \ln \left| \mbf{Q}^{-1} \right| - \frac{1}{2} {P} \ln \left| \mbf{R}^{-1} \right|, \\
% V_2
V_2 =& \frac{1}{2} {P} \lambda_A \left\| \mbf{A} \right\|^2_{\mbf{Q}^{-1}}  + \frac{1}{2} {P} \lambda_B \left\| \mbf{B} \right\|^2_{\mbf{Q}^{-1}}  +  \frac{1}{2} {P} \lambda_H \left\| \mbf{H} \right\|^2_{\mbf{Q}^{-1}} \nonumber \\
&+  \frac{1}{2} {P} \lambda_C \left\| \mbf{C} \right\|^2_{\mbf{R}^{-1}} + \frac{1}{2} {P} \lambda_Q \, \mbox{tr}(\mbf{Q}^{-1}) +  \frac{1}{2} {P} \lambda_R \, \mbox{tr}(\mbf{R}^{-1}).
\end{align}
\end{subequations}
Here, the norm is a weighted Frobenius matrix norm: $\left\| \mbf{X} \right\|_{\mbf{W}} = \sqrt{\mbox{tr}\left( \mbf{X}^T \mbf{W} \mbf{X} \right)}$. $V_1$ represents the negative log-likelihood of the Bayesian posterior from fitting the data, ignoring the normalizing constant. $V_2$ are prior terms over the matrices, where the first four terms encourage the description length of $\mbf{A}$, $\mbf{B}$, $\mbf{H}$, and $\mbf{C}$ to be minimal while the last two are (isotropic) inverse-Wishart (IW) priors for the covariances $\mbf{Q}$ and $\mbf{R}$. IW distributions have been demonstrated to be robust priors for learning covariances \cite{esgvi-extended}. The regularizing hyperparameters, $\lambda_A,\lambda_B,\lambda_H,\lambda_C,\lambda_Q,\lambda_R$, will be later tuned according to the data gathered.

We find the critical points by setting derivatives of $V$ with respect to the model parameters $(\frac{\partial V}{\partial \mbf{A}}$, $\frac{\partial V}{\partial \mbf{B}}$, $\frac{\partial V}{\partial \mbf{H}}$, $\frac{\partial V}{\partial \mbf{C}}$, $\frac{\partial V}{\partial \mbf{Q}^{-1}}$, and $\frac{\partial V}{\partial \mbf{R}^{-1}})$ to zero. We define
\begin{gather}
 \mbf{V} = \mbf{U} \odot \tilde{\mbf{X}}, \quad 
 \mbf{J}= \mbf{X} - \mbf{A} \tilde{\mbf{X}} - \mbf{B} \mbf{U} - \mbf{H}\mbf{V}.
\end{gather}
This yields the following expressions:
\begin{subequations}\label{eq:abhcqr}
\begin{gather}
\begin{bmatrix}
\tilde{\mbf{X}}\tilde{\mbf{X}}^T + {P} \lambda_A \mbf{1} &
\tilde{\mbf{X}}\mbf{U}^T &
\tilde{\mbf{X}}\mbf{V}^T \\
\mbf{U}\tilde{\mbf{X}}^T &
\mbf{U}\mbf{U}^T + {P} \lambda_B \mbf{1} &
\mbf{U} \mbf{V}^T \\
\mbf{V}\tilde{\mbf{X}}^T &
\mbf{V} \mbf{U}^T &
\mbf{V}\mbf{V}^T + {P} \lambda_H \mbf{1}
\end{bmatrix}
\begin{bmatrix} \mbf{A}^T \\ \mbf{B}^T \\ \mbf{H}^T \end{bmatrix}
=
\begin{bmatrix} \tilde{\mbf{X}}\mbf{X}^T \\ \mbf{U}\mbf{X}^T \\ \mbf{V}\mbf{X}^T \end{bmatrix},
\end{gather}
\begin{align}
\mbf{C} & = (\mbf{Y}\mbf{X}^T) (\mbf{X}\mbf{X}^T + {P} \lambda_C \mbf{1})^{-1}, \\
\mbf{Q} & = \frac{1}{{P}} \mbf{J} \mbf{J}^T + \lambda_A \mbf{A} \mbf{A}^T + \lambda_B \mbf{B} \mbf{B}^T + \lambda_H \mbf{H} \mbf{H}^T + \lambda_Q \mbf{1} , \\
\mbf{R} & = \frac{1}{{P}} (\mbf{Y} - \mbf{C} \mbf{X})(\mbf{Y} - \mbf{C} \mbf{X})^T + \lambda_C \mbf{C} \mbf{C}^T +  \lambda_R \mbf{1} ,
\end{align}
\end{subequations}
where $\mbf{1}$ represents the identity operator for the appropriate domains. We can solve for $\mbf{A}$, $\mbf{B}$, and $\mbf{H}$ through solving a system of linear equations, then use these results to find $\mbf{Q}$ and $\mbf{R}$. This procedure is linear in the amount of training data, $P$, for both computation and storage.

\section{Batch Linear State Estimation}
\label{sec:batch-est}
Having learned the system matrices from training data, we now wish to solve for a sequence of test states, $\{\mbs{\xi}_k^\prime\}_{k=0}^K$, given a series of inputs, $\{\mbs{\nu}_k^\prime\}_{k=1}^K$, and measurements, $\{\mbs{\gamma}_k^\prime\}_{k=0}^K$, where $(\cdot)^\prime$ denotes quantities at test time. We do this in the lifted space, where the quantities become $\{\mbf{x}_k^\prime\}_{k=0}^K$, $\{\mbf{u}_k^\prime\}_{k=1}^K$, and $\{\mbf{y}_k^\prime\}_{k=0}^K$, respectively. The main insight is that since the inputs are completely determined at test time, we can manipulate the lifted time-invariant bilinear form of \eqref{eq:bilin-stochastic} into a lifted time-varying linear form. We observe that
\begin{equation}
 \mbf{u}_k^\prime \otimes \mbf{x}_{k-1}^\prime = (\mbf{u}_k^\prime \otimes \mbf{1}) \mbf{x}_{k-1}^\prime,
\end{equation}
where here $\mbf{1} \col \mathcal{X} \to \mathcal{X}$. The motion model for the test trajectory becomes, for $k=1,\dots,K$,
\begin{gather}
 \mbf{x}_{k}^\prime = \mbf{A}\mbf{x}_{k-1}^\prime + \mbf{B}\mbf{u}_k^\prime + \mbf{H} \left(\mbf{u}_k^\prime \otimes \mbf{1}\right) \mbf{x}_{k-1}^\prime + \mbf{w}_k^\prime.
\end{gather}
As $\mbf{u}_k^\prime$ is given at test time, we define a new time-varying system matrix $\mbf{A}_{k-1}$ and input $\mbf{v}_k^\prime$ as
\begin{equation}
\label{eq:convert-to-ltv}
  \mbf{A}_{k-1} = \mbf{A} + \mbf{H}(\mbf{u}_k^\prime \otimes \mbf{1}), \quad \mbf{v}_k^\prime = \mbf{B}\mbf{u}_k^\prime.
\end{equation}
With this, we have converted the bilinear system into a linear time-varying (LTV) system, governed by
\pagebreak
\begin{subequations}\label{eq:ltv-system}
\begin{align}
 \mbf{x}_{k}^\prime & = \mbf{A}_{k-1} \mbf{x}_{k-1}^\prime + \mbf{v}_k^\prime + \mbf{w}_k^\prime, \quad &k=1,\dots,K, \\
 \mbf{y}_k^\prime & = \mbf{C} \mbf{x}_k^\prime + \mbf{n}_k^\prime, \quad &k=0,\dots,K,
\end{align}
\end{subequations}
where $\mbf{w}_k^\prime \sim \mathcal{N}(\mbf{0},\mbf{Q})$ and $\mbf{n}_k^\prime \sim \mathcal{N}(\mbf{0},\mbf{R})$. This is the well-established batch state-estimation problem on linear-Gaussian systems \cite{barfoot-txtbk}. The solution is in the form of
\begin{equation}
 \mbf{x}^\prime_k \sim \mathcal{N}(\hat{\mbf{x}}^\prime_k, \hat{\mbf{P}}^\prime_k), \quad k = 0,\dots,K.
\end{equation}
$\hat{\mbf{x}}^\prime_k$ and $\hat{\mbf{P}}^\prime_k$ are, respectively, the mean and covariance estimates for the test trajectory. One popular method for solving \eqref{eq:ltv-system} is the Rauch-Tung-Striebel (RTS) smoother, but there are other efficient methods for obtaining exact solutions \cite{barfoot-txtbk}.

\section{Recovering Estimates from Lifted Space}
\label{sec:recover_estimates}
Having solved for the state estimates and covariances in the RKHS, we now wish to recover these quantities in the original space. Using the Representer Theorem \cite{representer-thm}, since the solution of the LTV system in \eqref{eq:ltv-system} is the result of a linear optimization problem, it must be spanned by the training data used to form the system matrices. We can thus write the state and covariance outputs of each timestep as
\begin{align}
 \hat{\mbf{x}}_k^\prime = \mbf{X} \tilde{\mbf{x}}_k, \quad
 \hat{\mbf{P}}_k^\prime = \mbf{X} \tilde{\mbf{P}}_k \mbf{X}^T,
\end{align}
where $\tilde{\mbf{x}}_k \in \mathbb{R}^{P}$, $\tilde{\mbf{P}}_k \in \mathbb{R}^{P\times P}$ consist of the appropriate weights, which we solve using the left pseudoinverse with a small hyperparameter $\lambda_x$ to regularize the inversion of $\mbf{X}^T \mbf{X}$:
\begin{subequations}
\begin{align}
 \tilde{\mbf{x}}_k &\approx (\mbf{X}^T \mbf{X} + \lambda_x \mbf{1})^{-1} \mbf{X}^T \hat{\mbf{x}}^\prime_k, \\
 \tilde{\mbf{P}}_k &\approx (\mbf{X}^T \mbf{X} + \lambda_x \mbf{1})^{-1} \mbf{X}^T \hat{\mbf{P}}^\prime_k \mbf{X}(\mbf{X}^T \mbf{X} + \lambda_x \mbf{1})^{-1}.
\end{align}
\end{subequations}
Now, by the properties of the mean map and the covariance map, these same weights can be used to construct the mean states and covariances in the original space through the weighted linear combination of training data,
\begin{equation}
 \hat{\mbs{\xi}}_k^\prime = \mbs{\Xi} \tilde{\mbf{x}}_k, \quad \hat{\mbs{\Sigma}}_k^\prime = \mbs{\Xi} \tilde{\mbf{P}}_k \mbs{\Xi}^T,
\end{equation}
where $\mbs{\xi}_k^\prime \sim \mathcal{N}(\hat{\mbs{\xi}}_k^\prime, \hat{\mbs{\Sigma}}_k^\prime)$ is the state at timestep $k$. We can then solve for the state means and covariances with
\begin{subequations}\label{eq:outputs}
\begin{align}
 \hat{\mbs{\xi}}_k^\prime &= \mbs{\Xi} (\mbf{X}^T \mbf{X} + \lambda_x \mbf{1})^{-1} \mbf{X}^T \hat{\mbf{x}}^\prime_k, \\
 \hat{\mbs{\Sigma}}_k^\prime &= \mbs{\Xi} (\mbf{X}^T \mbf{X} + \lambda_x \mbf{1})^{-1} \mbf{X}^T \hat{\mbf{P}}^\prime_k \mbf{X}(\mbf{X}^T \mbf{X} + \lambda_x \mbf{1})^{-1} \mbs{\Xi}^T.
\end{align}
\end{subequations}
To facilitate efficient computation when using RFFs in Section \ref{sec:approx-with-rff}, we define $\mbf{O}_{\mbs{\xi}} \col \mathcal{X} \to N_\xi$ as $\mbf{O}_{\mbs{\xi}} = \mbs{\Xi} \mbf{X}^T (\mbf{X}\mbf{X}^T + \lambda_x \mbf{1})^{-1}$, which can be precomputed during training. Using Sherman-Morrison-Woodbury (SMW) identities, \eqref{eq:outputs} becomes
\begin{align}\label{eq:outputs-short}
 \hat{\mbs{\xi}}_k^\prime = \mbf{O}_{\mbs{\xi}} \hat{\mbf{x}}_k^\prime, \quad \hat{\mbs{\Sigma}}_k^\prime = \mbf{O}_{\mbs{\xi}} \hat{\mbf{P}}_k^\prime \mbf{O}_{\mbs{\xi}}^T.
\end{align}
Note that \eqref{eq:outputs-short} assumes that the original states are within a vector space. The states could instead contains angles or other quantities on a circular domain. In this case, $\hat{\mbs{\xi}}_k^\prime$ should be a weighted average of circular quantities $\mbs{\Xi}$ with weights $\mbf{X}^T (\mbf{X}\mbf{X}^T + \lambda_x \mbf{1})^{-1} \hat{\mbf{x}}_i^\prime$ computed as usual, since the RKHS embeddings are within a vector space, and a similar weighted average is done for $\hat{\mbs{\Sigma}}_k^\prime$. We still use \eqref{eq:outputs-short} but slightly modify the computation of $\mbf{O}_{\mbs{\xi}}$. See Section \ref{sec:experiments} for an example.

\section{Approximate Embeddings with Random Fourier Features}
\label{sec:approx-with-rff}

So far, we have been working in the RKHS space and using the feature maps directly. For many kernels, however, these RKHS feature maps are very high (potentially infinite) dimensional. As it is often unclear how to truncate the feature maps directly to get reasonable approximations, we look for another form of approximate embeddings for $\mbf{x}(\cdot)$, $\mbf{u}(\cdot)$, and $\mbf{y}(\cdot)$ such that the solution computed through the algorithm approaches that from the true embeddings.

We notice that all of the RKHS quantities computed in our procedure appear only to be involved in the form of inner products, $\mbf{X}^T\mbf{X}$, or outer products, $\mbf{X}\mbf{X}^T$. Although the inner product is closely related to kernels, we look to avoid using only kernel evaluations for computation as this would yield a cost of at least $\mathcal{O}(P^3)$. Thus, we turn to Random Fourier Features (RFF) \cite{rff} for deriving approximate embeddings. Suppose we have two quantities, $\mbs{\xi}_i$ and $\mbs{\xi}_j$, in some state space, an operator, $\mbf{x}(\cdot)$, for lifting them into RKHS embeddings, $\mbf{x}_i$ and $\mbf{x}_j$, and the kernel function, $\kappa$, associated with this RKHS. We can use $\breve{\mbf{x}}(\cdot)$, the RFF embedding corresponding to $\kappa$, to approximate the kernel evaluation as
\begin{equation}
 \mbf{x}_i^T \mbf{x}_j = \mbf{x}(\mbs{\xi}_i)^T \mbf{x}(\mbs{\xi}_j) = \kappa(\mbs{\xi}_i, \mbs{\xi}_j) \approx \breve{\mbf{x}}(\mbs{\xi}_i)^T \breve{\mbf{x}}(\mbs{\xi}_j).
\end{equation}
We see that as long as any RKHS quantities within an expression are involved in the form of inner products within the same space (i.e., kernelized), we can swap the exact RKHS embeddings with their RFF embeddings for a valid approximation. We can freely manipulate RKHS quantities within kernelized expressions for efficient computation, and the approximation from swapping the embeddings from RKHS to RFF would still be valid.

\subsection{Sketch that KoopSE is Kernelized}
\label{sec:kernel-proof}
Since we look to use RFF as approximate embeddings for $\mbf{x}(\cdot)$, $\mbf{u}(\cdot)$, and $\mbf{y}(\cdot)$, our goal is to show that KoopSE uses the RKHS quantities only in their kernelized forms, even if these forms are not actually used for computation. See Appendix \ref{sec:kernel-proof-appendix} for additional details of the proof sketch.
  % see top level paper for def of this command

 For training, the lifted training points are used to compute the bilinear system matrices in \eqref{eq:bilin-stochastic}. Using SMW identities, it can be seen that the analytical solution of \eqref{eq:abhcqr} has the form
\begin{subequations}
\begin{alignat}{2}
\mbf{A} & = \mbf{X} \mbf{W}_A \mbf{X}^T, \quad
&\mbf{B} &= \mbf{X} \mbf{W}_B \mbf{U}^T, \\
\mbf{H} & = \mbf{X} \mbf{W}_H \mbf{V}^T, \quad
&\mbf{C} &= \mbf{Y} \mbf{W}_C \mbf{X}^T, \\
 \mbf{Q} & = \mbf{X} \mbf{W}_Q \mbf{X}^T + \lambda_Q \mbf{1}, \quad
&\mbf{R} &= \mbf{Y} \mbf{W}_R \mbf{Y}^T + \lambda_R \mbf{1},
\end{alignat}
\end{subequations}
where $\mbf{W}_A, \mbf{W}_B, \mbf{W}_H, \mbf{W}_C, \mbf{W}_Q, \mbf{W}_R$ are some kernelized matrices. At test time, we assume that the initial condition, new control inputs, and new measurements can be written as linear combinations of those seen in training:
\begin{subequations}
 \begin{align}
  \check{\mbf{x}}^\prime_0 &= \mbf{X}\mbf{w}_{\check{x},0}, \\
  {\mbf{u}}^\prime_k &= \mbf{U}\mbf{w}_{u,k}, \quad k = 1,\dots,K \\
  {\mbf{y}}^\prime_k &= \mbf{Y}\mbf{w}_{y,k}, \quad k = 0,\dots,K
 \end{align}
 \end{subequations}
for weights $\mbf{w}_{\check{x},0}$, $\mbf{w}_{u,k}$, and $\mbf{w}_{y,k}$. Then, it can be seen from \eqref{eq:convert-to-ltv} that the time-varying quantities have the form
\begin{align}
 \mbf{A}_{k-1} = \mbf{X}\mbf{W}_{A,k}\mbf{X}^T, \quad \mbf{v}_k^\prime = \mbf{X}\mbf{w}_{v,k}, \quad k=0,\dots,K
\end{align}
for some kernelized matrices, $\mbf{W}_{A,k}$ and $\mbf{w}_{v,k}$. Now, the solution to \eqref{eq:ltv-system} is exactly solved by the RTS smoother \cite{barfoot-txtbk}. It is then straightforward to prove through two induction proofs, one for the forward pass and one for the backward pass, that the state means and covariances for both passes have the forms $\hat{\mbf{x}}_k^\prime = \mbf{X}\mbf{w}_{\hat{x},k}$, $
 \hat{\mbf{P}}_k^\prime = \mbf{X}\mbf{W}_{\hat{P},k}\mbf{X}^T + c_k\mbf{1}$, for $k=0,\dots,K$ and some kernelized matrices $\mbf{w}_{\hat{x},k}, \mbf{W}_{\hat{P},k}$ and scalar $c_k$. By substituting these expressions into \eqref{eq:outputs}, it can be seen that the results for $\hat{\mbs{\xi}}_k^\prime$ and $\hat{\mbs{\Sigma}}^\prime_k$, the means and covariances in the original space, are indeed kernelized. Therefore, from input to output, the algorithm only uses RKHS quantities in their kernelized forms.

\subsection{Substituting with Random Fourier Features}
As the algorithm is kernelized, we can approximate the kernel functions associated with embeddings $\mbf{x}(\cdot), \mbf{y}(\cdot), \mbf{u}(\cdot)$ by directly swapping them with the respective finite-dimensional RFF embeddings on the original quantities. When $P$ is large, this yields a much more efficient procedure than using only kernel evaluations. We outline the procedure below, and we analyze the time and memory complexity of the algorithm.

Let $\breve{\mbf{x}}:\mathbb{R}^{N_\xi}\rightarrow\mathbb{R}^{R_x}$, $\breve{\mbf{u}}:\mathbb{R}^{N_\nu}\rightarrow\mathbb{R}^{R_u}$, and $\breve{\mbf{y}}:\mathbb{R}^{N_\gamma}\rightarrow\mathbb{R}^{R_y}$ represent the RFF embedding for $\mbs{\xi}$, $\mbs{\nu}$, and $\mbs{\gamma}$, with $R_x,R_u,R_y$ being the respective ranks of the approximation. We replace the RKHS embeddings with their RFF counterparts, denoted by $\breve{(\cdot)}$, for the training quantities:
\begin{gather}
 \mbf{x}_i \leftarrow \breve{\mbf{x}}_i = \breve{\mbf{x}}(\mbs{\xi}_i), \quad
 \mbf{u}_i \leftarrow \breve{\mbf{u}}_i = \breve{\mbf{u}}(\mbs{\nu}_i), \quad
 \mbf{y}_i \leftarrow \breve{\mbf{y}}_i = \breve{\mbf{y}}(\mbs{\eta}_i),
\end{gather}
where $i = 1,\dots,P$. The embedded training data in block-matrix form becomes
\begin{subequations}
\begin{gather}
 \mbf{X} \leftarrow \breve{\mbf{X}} \in \mathbb{R}^{R_x \times P}, \quad
 \tilde{\mbf{X}} \leftarrow \breve{\tilde{\mbf{X}}} \in \mathbb{R}^{R_x \times P}, \\
 \mbf{Y} \leftarrow \breve{\mbf{Y}} \in \mathbb{R}^{R_y \times P}, \quad
 \mbf{U} \leftarrow \breve{\mbf{U}} \in \mathbb{R}^{R_u \times P},
 \end{gather}
 \end{subequations}
 which are used to compute the now finite-dimensional RFF-counterpart of the system matrices,
 \begin{subequations}
 \begin{gather}
  \breve{\mbf{A}}\in\mathbb{R}^{R_x\times R_x}, \;
  \breve{\mbf{B}}\in\mathbb{R}^{R_x\times R_u}, \;
  \breve{\mbf{H}}\in\mathbb{R}^{R_x\times (R_x R_u)}, \\
  \breve{\mbf{C}}\in\mathbb{R}^{R_y\times R_x}, \; 
  \breve{\mbf{Q}}\in\mathbb{R}^{R_x\times R_x}, \; 
  \breve{\mbf{R}}\in\mathbb{R}^{R_y\times R_y},
 \end{gather}
 \end{subequations}
through solving \eqref{eq:abhcqr} with the embedded training data. Letting $R=\max(R_x,R_u,R_y)$, this procedure has a computational complexity of $\mathcal{O}(PR^3)$ and a memory complexity of $\mathcal{O}(PR^2)$, both scaling only linearly with the number of training points.
 
For testing, we replace the initial condition, incoming control inputs, and incoming measurements with their RFFs,
 \begin{subequations}
 \begin{align}
   \check{\mbf{x}}_0^\prime &\leftarrow \breve{\mbf{x}}(\check{\mbs{\xi}}_0^\prime) = \breve{\check{\mbf{x}}}_0^\prime, \\
   \mbf{u}_k^\prime &\leftarrow \breve{\mbf{u}}(\mbs{\nu}_k^\prime) = \breve{\mbf{u}}_k^\prime, \quad k = 1,\dots,K \\
   \mbf{y}_k^\prime &\leftarrow \breve{\mbf{y}}(\mbs{\eta}_k^\prime) = \breve{\mbf{y}}_k^\prime, \quad k = 0,\dots,K
 \end{align}
 \end{subequations}
and we form $\breve{\mbf{A}}_{k-1} \in \mathbb{R}^{R_x\times R_x}$, $\breve{\mbf{v}}_k \in \mathbb{R}^{R_x}$ using \eqref{eq:convert-to-ltv} with the $\breve{(\cdot)}$ quantities. We then solve the RFF-equivalent of the LTV system in \eqref{eq:ltv-system} using a linear batch state estimator, yielding the RFF-equivalent state mean, $\breve{\hat{\mbf{x}}}_k^\prime$, and covariance, $\breve{\hat{\mbf{P}}}_k^\prime$, for each timestep $k=0,\dots,K$. With an RTS smoother or a similarly efficient estimator, this procedure takes a computation complexity of $\mathcal{O}(KR^3)$ and a memory complexity of $\mathcal{O}(KR^2)$. Note that if we require a filter solution for online state estimation, we can do only the forward pass, which is also kernelized.

Finally, we convert the results back into state space:
\begin{subequations}\label{eq:outputs-rff}
\begin{gather}
 \hat{\mbs{\xi}}_k^\prime = \breve{\mbf{O}}_{\mbs{\xi}} \breve{\hat{\mbf{x}}}_k^\prime, \quad
 \hat{\mbs{\Sigma}}_k^\prime = \breve{\mbf{O}}_{\mbs{\xi}} \breve{\hat{\mbf{P}}}^\prime_k \breve{\mbf{O}}_{\mbs{\xi}}^T, \label{eq:outputs-rff-1} \\
 \breve{\mbf{O}}_{\mbs{\xi}} = \mbs{\Xi} \breve{\mbf{X}}^T (\breve{\mbf{X}}\breve{\mbf{X}}^T + \lambda_x \mbf{1})^{-1} \in \mathbb{R}^{N_\xi \times R_x}, \label{eq:outputs-rff-2}
\end{gather}
\end{subequations}
where $k=0,\dots,K$. Due to kernelization, although the RFF-equivalents of the RKHS variables and model matrices likely look very different, the results for $\{\hat{\mbs{\xi}}_k^\prime,\hat{\mbs{\Sigma}}_k^\prime\}$ in \eqref{eq:outputs-rff} converge to the true results in \eqref{eq:outputs-short} as the ranks of the approximations approaches infinity. Note that $\breve{\mbf{O}}_{\mbs{\xi}}$ can be precomputed during training. As such, the overall computation and memory complexity of testing is still $\mathcal{O}(KR^3)$ and $\mathcal{O}(KR^2)$, respectively, regardless of the amount of training data used. As mentioned before, the computation of $\breve{\mbf{O}}_{\mbs{\xi}}$ would need to be slightly modified if there were circular quantities in the original states. See Algorithm \ref{alg-summary} for a summary of the full algorithm.

\begin{figure}[!t]
\removelatexerror
\begin{algorithm}[H]
\small
\SetAlgoLined
\DontPrintSemicolon
\hskip -1.0em \textbf{Training:} \\
 \KwIn{Training data $\{ \tilde{\mbs{\xi}}^{(i)}, \mbs{\xi}^{(i)}, \mbs{\nu}^{(i)}, \mbs{\gamma}^{(i)} \}_{i=1}^P$; RFF parameters (kernel-dependent); algorithm hyperparameters $\{\lambda_A, \lambda_B, \lambda_H, \lambda_C, \lambda_Q, \lambda_R, \lambda_x\}$.}
\begin{enumerate}
 \item
 Stack training data $\{ \tilde{\mbs{\xi}}^{(i)}, \mbs{\xi}^{(i)}, \mbs{\nu}^{(i)}, \mbs{\gamma}^{(i)} \}_{i=1}^P$ into their block-matrix form, $\{\tilde{\mbs{\Xi}}, \mbs{\Xi}, \mbs{N}, \mbs{\Gamma}\}$, with \eqref{eq:block-matrix-train}.
 \item 
 Embed $\{\tilde{\mbs{\Xi}}, \mbs{\Xi}, \mbs{N}, \mbs{\Gamma}\}$ into their RFF embeddings in block-matrix form, yielding $\{\breve{\tilde{\mbf{X}}}, \breve{\mbf{X}}, \breve{\mbf{U}}, \breve{\mbf{Y}}\}$.
 \item
 Solve for model matrices $\{\breve{\mbf{A}}, \breve{\mbf{B}}, \breve{\mbf{H}}, \breve{\mbf{C}}, \breve{\mbf{Q}}, \breve{\mbf{R}}\}$ using \eqref{eq:abhcqr} \\ with the $\breve{(\cdot)}$ versions of training data, and compute $\breve{\mbf{O}}_{\mbs{\xi}}$ \\ in \eqref{eq:outputs-rff-2}.
\end{enumerate}
 \KwOut{$\breve{\mbf{A}}, \breve{\mbf{B}}, \breve{\mbf{H}}, \breve{\mbf{C}}, \breve{\mbf{Q}}, \breve{\mbf{R}}, \breve{\mbf{O}}_{\mbs{\xi}}$.}
\;
\hskip -1.0em \textbf{Testing:} \\
\KwIn{Initial condition $\check{\mbs{\xi}}_0$; control inputs $\{\mbs{\nu}^\prime_k\}_{k=1}^K$; measurements $\{\mbs{\gamma}^\prime_k\}_{k=0}^K$.}
\begin{enumerate}
 \item
 Embed $\left\{\check{\mbs{\xi}}_0^\prime, \{\mbs{\nu}^\prime_k\}_{k=1}^K, \{\mbs{\gamma}^\prime_k\}_{k=0}^K\right\}$ into their RFF \\ embeddings, yielding $\left\{\breve{\check{\mbf{x}}}_0^\prime, \{\breve{\mbf{u}}^\prime_k\}_{k=1}^K, \{\breve{\mbf{y}}^\prime_k\}_{k=0}^K\right\}$.
 \item
 Compute time-varying quantities $\{\breve{\mbf{A}}_{k-1},\breve{\mbf{v}}_k^\prime\}_{k=0}^K$ in \eqref{eq:convert-to-ltv} \\using the $\breve{(\cdot)}$ quantities.
 \item
 Form LTV system in \eqref{eq:ltv-system} with the $\breve{(\cdot)}$ quantities.
 \item
 Solve for mean and covariance estimates, $\{\breve{\hat{\mbf{x}}}_k^\prime, \breve{\hat{\mbf{P}}}_k^\prime\}_{k=0}^K$, \\ with an efficient batch state estimator (e.g., RTS smoother).
 \item
 Convert $\{\breve{\hat{\mbf{x}}}_k^\prime, \breve{\hat{\mbf{P}}}_k^\prime\}_{k=0}^K$ into the original space with \eqref{eq:outputs-rff-1}, yielding $\{\hat{\mbs{\xi}}_k^\prime, \hat{\mbf{\Sigma}}_k^\prime\}_{k=0}^K$.
 \end{enumerate}
 \KwOut{State means $\{\hat{\mbs{\xi}}_k^\prime\}_{k=0}^K$; covariances $\{\hat{\mbf{\Sigma}}_k^\prime\}_{k=0}^K$.}

 \caption{Koopman State Estimator (KoopSE)}
 \label{alg-summary}
\end{algorithm}
\vspace{-5pt}
\end{figure}

\vspace{-2pt}
\section{Experiments and Results}
\label{sec:experiments}
\vspace{-2pt}
\subsection{Problem Setup}
KoopSE was evaluated on a control-affine system with a nonlinear measurement model in simulation, then on a experimental dataset with a similar setup. The problem was estimating the positions and headings of a wheeled robot driving in a 2D plane and receiving range measurements from five ultra-wideband (UWB) sensors. For timestep $k$, the state, input, and measurement are, respectively, 
\begin{equation}
 \mbs{\xi}_k = \bbm x_k \\ y_k \\ \theta_k \ebm, \quad
 \mbs{\nu}_k = \bbm u_k \\ \omega_k \ebm, \quad
 \mbs{\gamma}_k = \bbm r_{k,1} \\ \vdots \\ r_{k,5} \ebm,
\end{equation}
where $(x_k,y_k)$ is the robot's position, $\theta_k$ is its orientation, $u_k$ is its linear velocity, $\omega_k$ is its angular velocity, and $r_{k,j}$ is the range measurement from the robot to the $j$th UWB sensor. This uses the common state estimation practice of using interoceptive measurments as inputs in the process model \cite{probabilistic-robotics}. This problem setup is identical to the 2D version of the setup in Section 8.2 of \cite{barfoot-txtbk}.%The motion model is $\mbs{\xi}_k = \mbs{\xi}_{k-1} + T \begin{bmatrix} \cos\theta_{k-1} & 0 \\ \sin\theta_{k-1} & 0 \\ 0 & 1 \end{bmatrix}$.

For the state, we used a product of a squared-exponential kernel for the robot's position and a periodic kernel for its orientation. The respective formulas for generating these RFFs can be found in \cite{rff} and \cite{rff-periodic}. The combined RFF for the state is thus the Cartesian product of the two RFF sets \cite{rff-periodic}. For the input, we kept it as is since there were no improvements from lifting it to higher dimensions, thereby using a linear kernel. We used the squared-exponential RFF for the measurement.

Since orientation is on a circular domain, we take special care in computing the weighted average of states in \eqref{eq:outputs-rff}. We convert the orientations to their Cartesian form,
\begin{equation}
 \mbs{\xi}^{\star(i)} = \mbs{\xi}^\star(\mbs{\xi}^{(i)})
 = \bbm x^{(i)} & y^{(i)} & \cos(\theta^{(i)}) & \sin(\theta^{(i)}) \ebm^T,
\end{equation}
then use $\mbs{\Xi}^\star = \bbm \mbs{\xi}^{\star(1)} & \cdots & \mbs{\xi}^{\star(P)} \ebm$ to compute $\breve{\mbf{O}}_{\mbs{\xi}^\star}$ in \eqref{eq:outputs-rff-2} 
  instead of using $\mbs{\Xi}$. When computing $\hat{\mbs{\xi}}_k^\prime$ and $ \hat{\mbs{\Sigma}}_k^\prime$ in \eqref{eq:outputs-rff-1} at test time, we first compute their Cartesian-form equivalents: $\hat{\mbs{\xi}}_k^{\prime\star} = \breve{\mbf{O}}_{\mbs{\xi}^\star} \breve{\hat{\mbf{x}}}_k^\prime, \;
 \hat{\mbs{\Sigma}}_k^{\prime\star} = \breve{\mbf{O}}_{\mbs{\xi}^\star} \breve{\hat{\mbf{P}}}^\prime_k \breve{\mbf{O}}_{\mbs{\xi}^\star}^T$, giving us the Gaussian distribution of $x_k$, $y_k$, $\cos\theta_k$, and $\sin\theta_k$. We then use the method of \cite{wang-13} to estimate the distribution of $\theta_k$ given the Gaussians of $\cos\theta_k$ and $\sin\theta_k$.

For evaluating estimation algorithms, we use the root-mean-squared-error (RMSE) and the Mahalanobis distance (scaled by the degrees of freedom) of the estimated trajectories. An accurate estimator has an RMSE close to $0$, and a consistent estimator has a Mahalanobis distance close to $1$. We tuned the RFF parameters and the regularizing hyperparameters for KoopSE accordingly for these objectives. We compare our results with a model-based Lie-group extended RTS smoother, an extension of the Lie-group extended Kalman Filter in \cite{barfoot-txtbk} Section 8.2.4. Its model covariances are tuned with the same objectives, including increasing the measurement covariances for the two noisy sensors. KoopSE was first verified in simulation, then validated on an experimental dataset of the same setup. We present our results for the two scenarios below.

\subsection{Simulation Results}
\label{sec:simulation}
\begin{table}[!t]
\centering
\begin{tabular}{|c|c|c|}
 \hline
  & KoopSE & Model-Based \\
 \hline
 Translation RMSE (m) & $\mbf{0.026}$ & $0.053$ \\
 \hline
 Orientation RMSE (rad) & $\mbf{0.026}$ & $0.034$ \\
 \hline
 Translation Maha. distance & $0.915$ & $1.104$ \\
 \hline
 Orientation Maha. distance & $0.777$ & $0.548$ \\
 \hline
 \end{tabular}
  \caption{RMSE and Mahalanobis distance for KoopSE and the model-based extended RTS smoother for 100 trajectories of 1000 timesteps in simulation. The Mahalanobis distances for both are fairly close to 1, signifying that both algorithms are properly tuned under the ideal simulation environment. However, KoopSE has lower RMSE than the model-based smoother for both translation and orientation.\vspace{0pt}}
 \label{tab:simu-errors}
\end{table}
\begin{figure}[!t]
\vspace{15pt}
 \centering
 \begin{widepage}
 \subcaptionbox{KoopSE Errors}{
 \includegraphics[width=0.47\columnwidth]{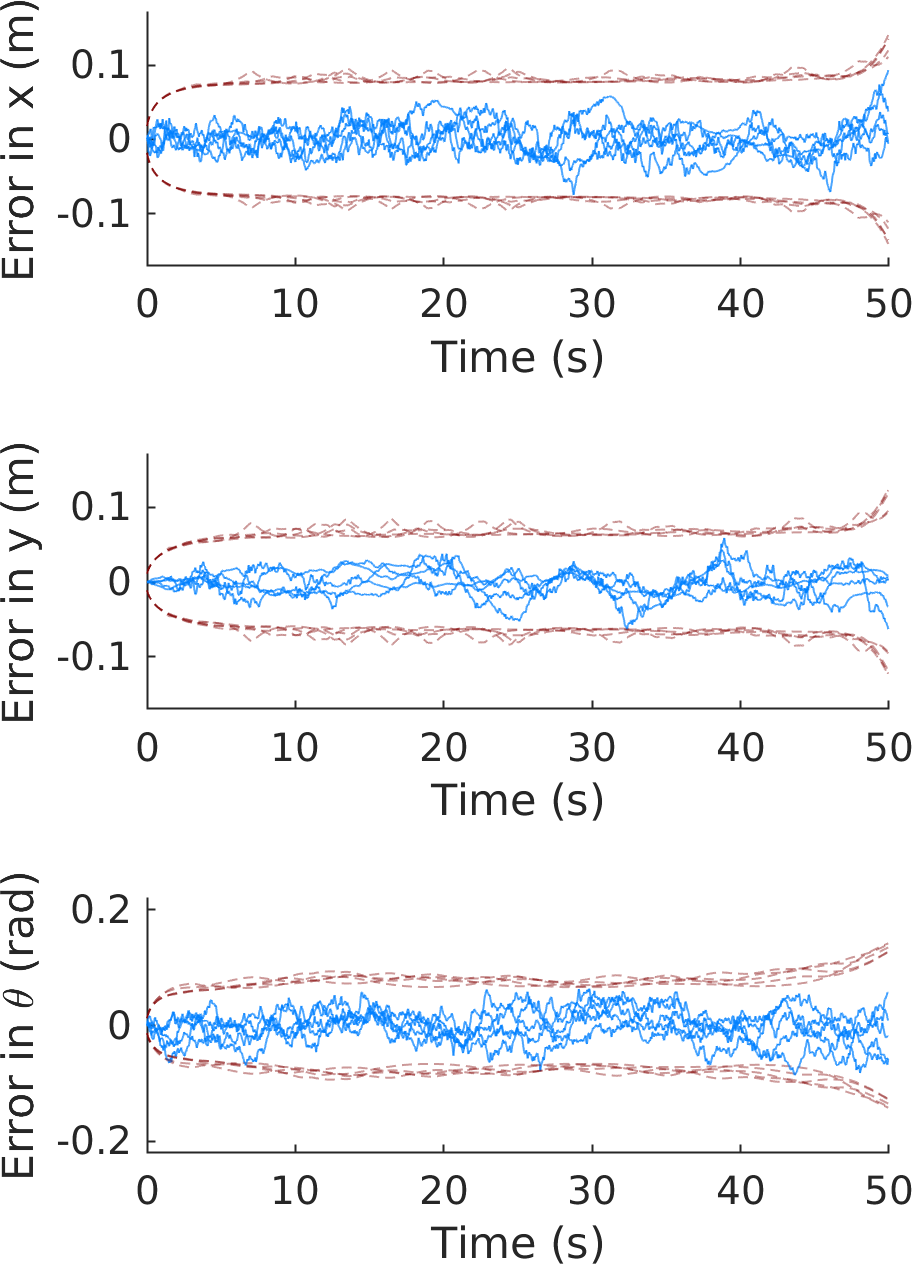}
 }
  \subcaptionbox{Model-Based Smoother Errors}{
 \includegraphics[width=0.47\columnwidth]{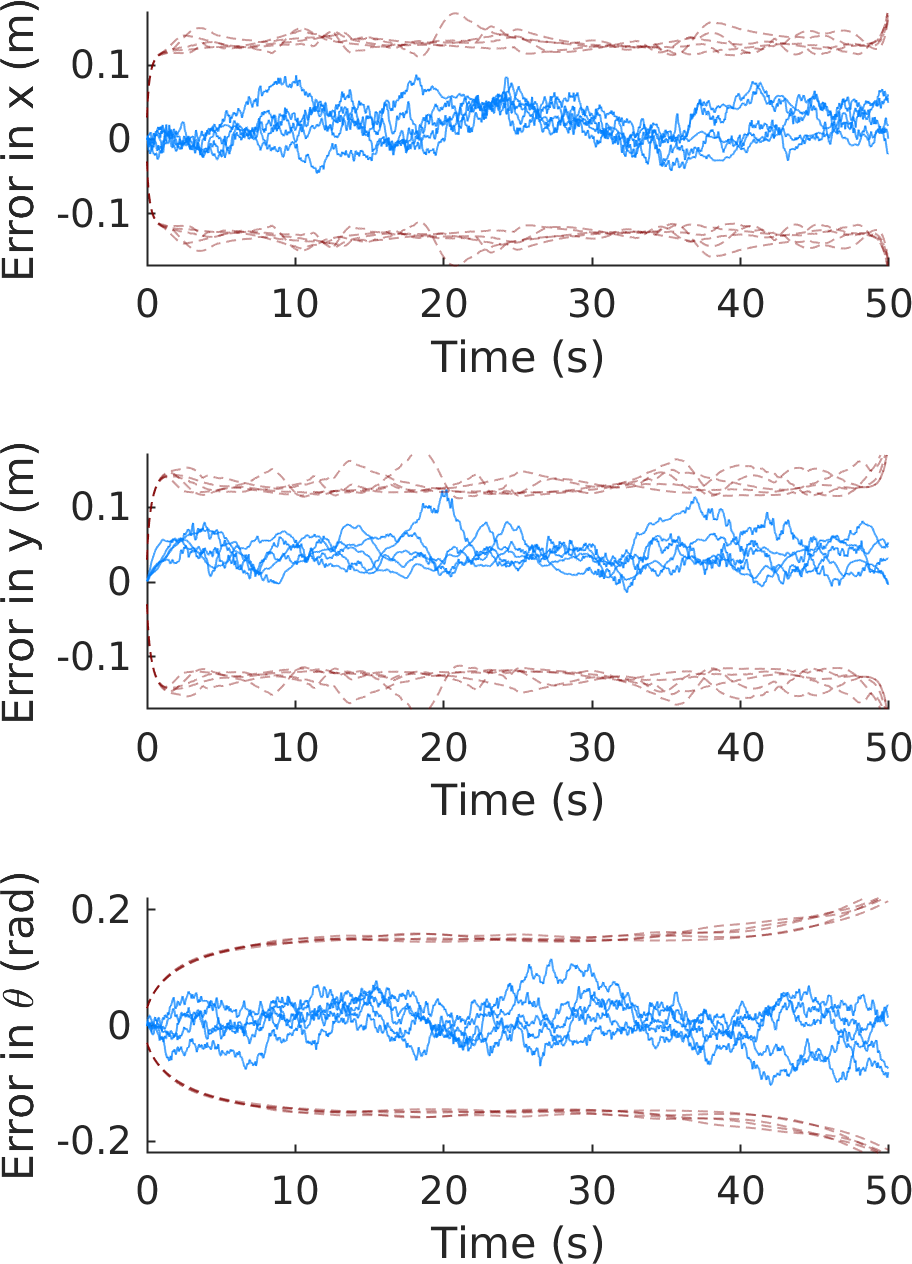}
 }
 \end{widepage}
 \caption{Error plots of 5 test trajectories in simulation for KoopSE (left) and for the model-based extended RTS smoother (right). The blue lines represent the errors of the estimated trajectories, and the red envelopes represent the estimated $3\sigma$ bounds. Both errors are within the $3\sigma$ bounds, but the model-based smoother has larger errors than KoopSE, especially for $y$ where we can see a small bias for the model-based smoother.\vspace{0pt}}
 \label{fig:simu-errors}
\end{figure}
\begin{figure}[!t]
\vspace{15pt}
 \centering
 \subcaptionbox*{}{
   \includegraphics[width=0.98\columnwidth]{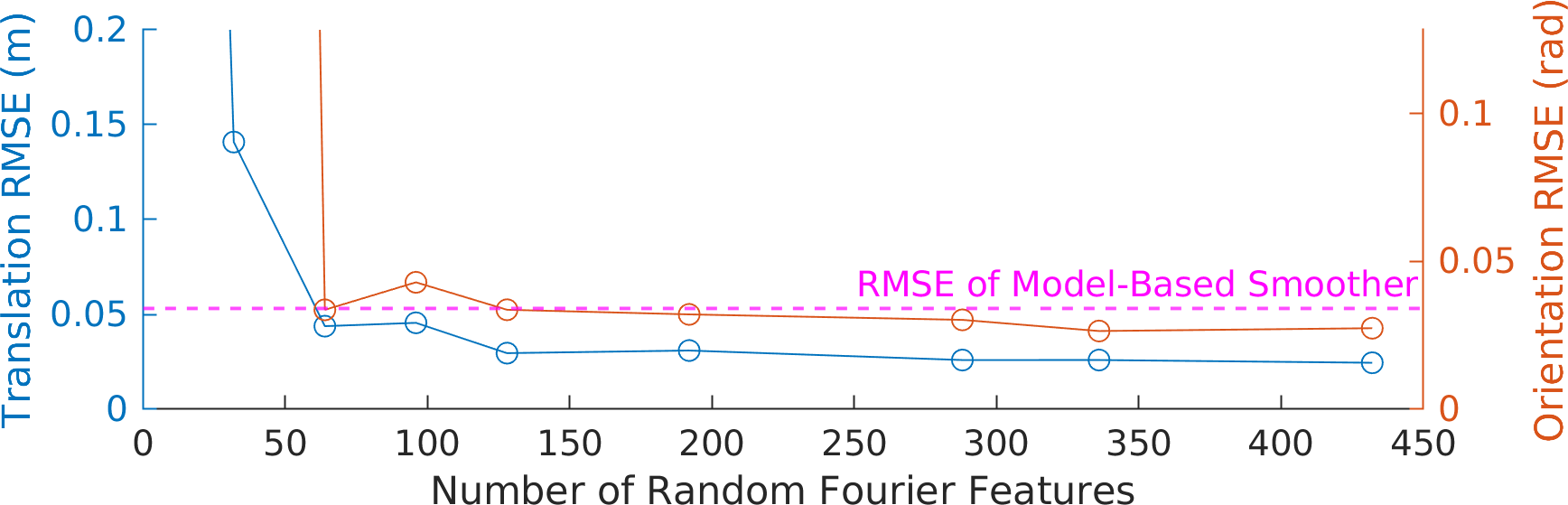}
   }
 \subcaptionbox*{}{
   \includegraphics[width=0.98\columnwidth]{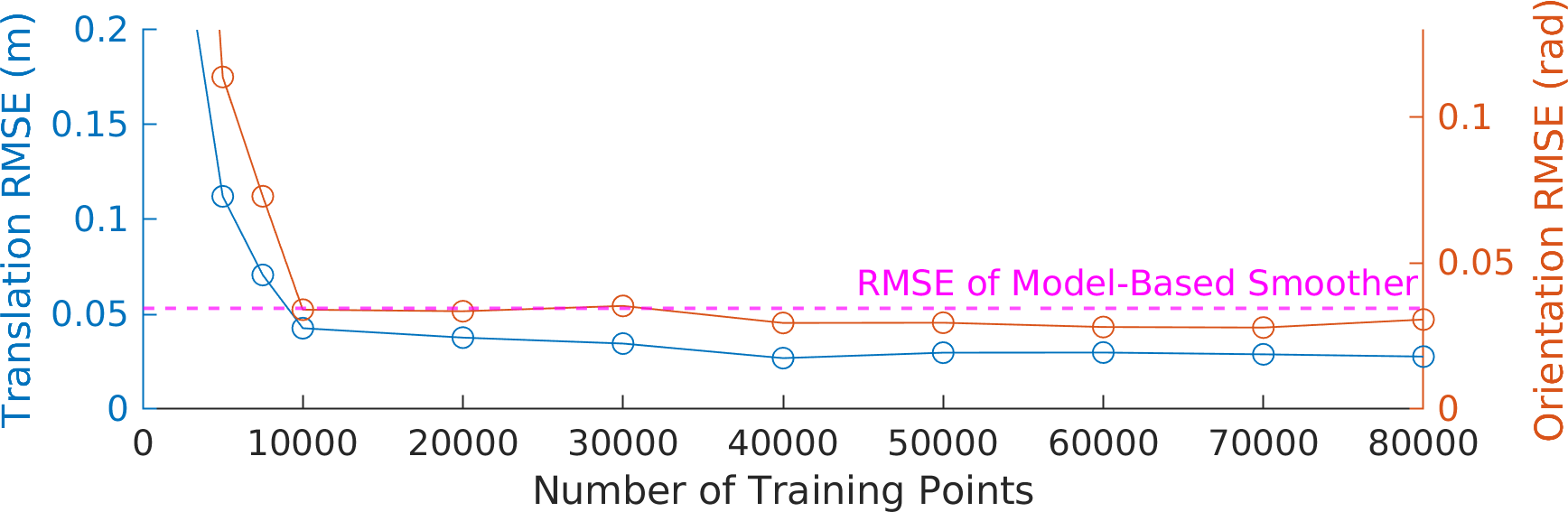}
}
\vspace{-10pt}
 \caption{RMSE of KoopSE on test trajectories in simulation for various numbers of RFF (top) and training data points (bottom). The dotted line represents the RMSE of the model-based smoother for both vertical axes (translation and orientation). Starting from using only 128 RFF and 10000 training points, the performance of KoopSE has already surpassed that of the model-based smoother, achieving much lower translation RMSE and comparable orientation RMSE.}
 \label{fig:simu-varying}
 \vspace{0pt}
\end{figure}
For the simulation, we added a $20$ cm unmodelled bias to two out of the five sensors, resulting in a UWB error profile similar to that gathered from experiment in Section \ref{sec:robot-exp}. Training data were generated by the robot roughly following randomly generated trajectories in an enclosed area. One hundred trajectories of 1000 timesteps were used for evaluation, and the combined results are presented in \tabref{tab:simu-errors}. The error plots for 5 trajectories are shown in \figref{fig:simu-errors}. In \figref{fig:simu-varying}, we present the results of testing KoopSE on the 5 trajectories using various numbers of RFF and number of training data points, in comparison to the model-based smoother.
\subsection{Experimental Results}
\label{sec:robot-exp}
\begin{figure}[!t]
 \centering
 \begin{widepage}
 \subcaptionbox{KoopSE Errors}{
 \includegraphics[width=0.48\columnwidth]{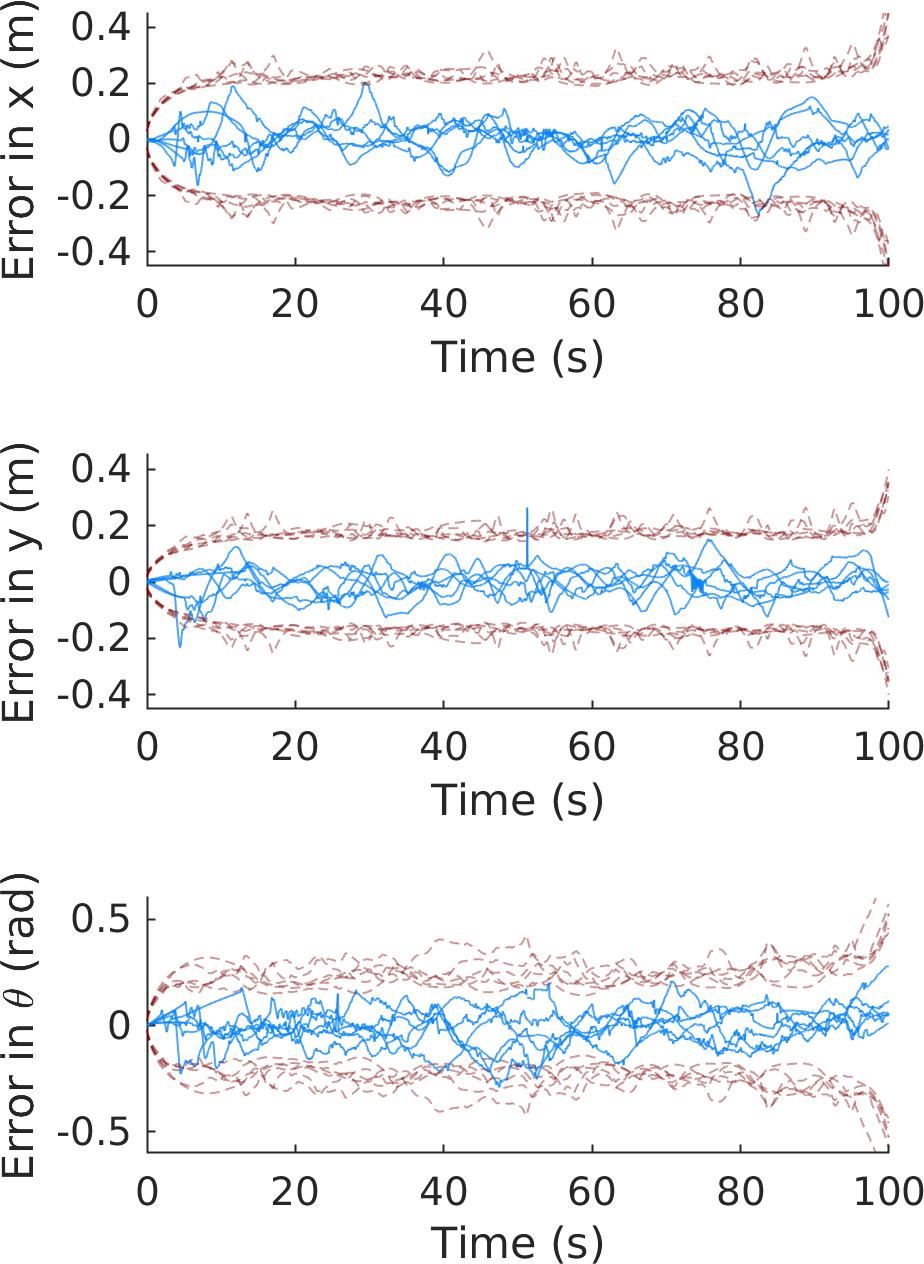}
 }
  \subcaptionbox{Model-Based Smoother Errors}{
 \includegraphics[width=0.48\columnwidth]{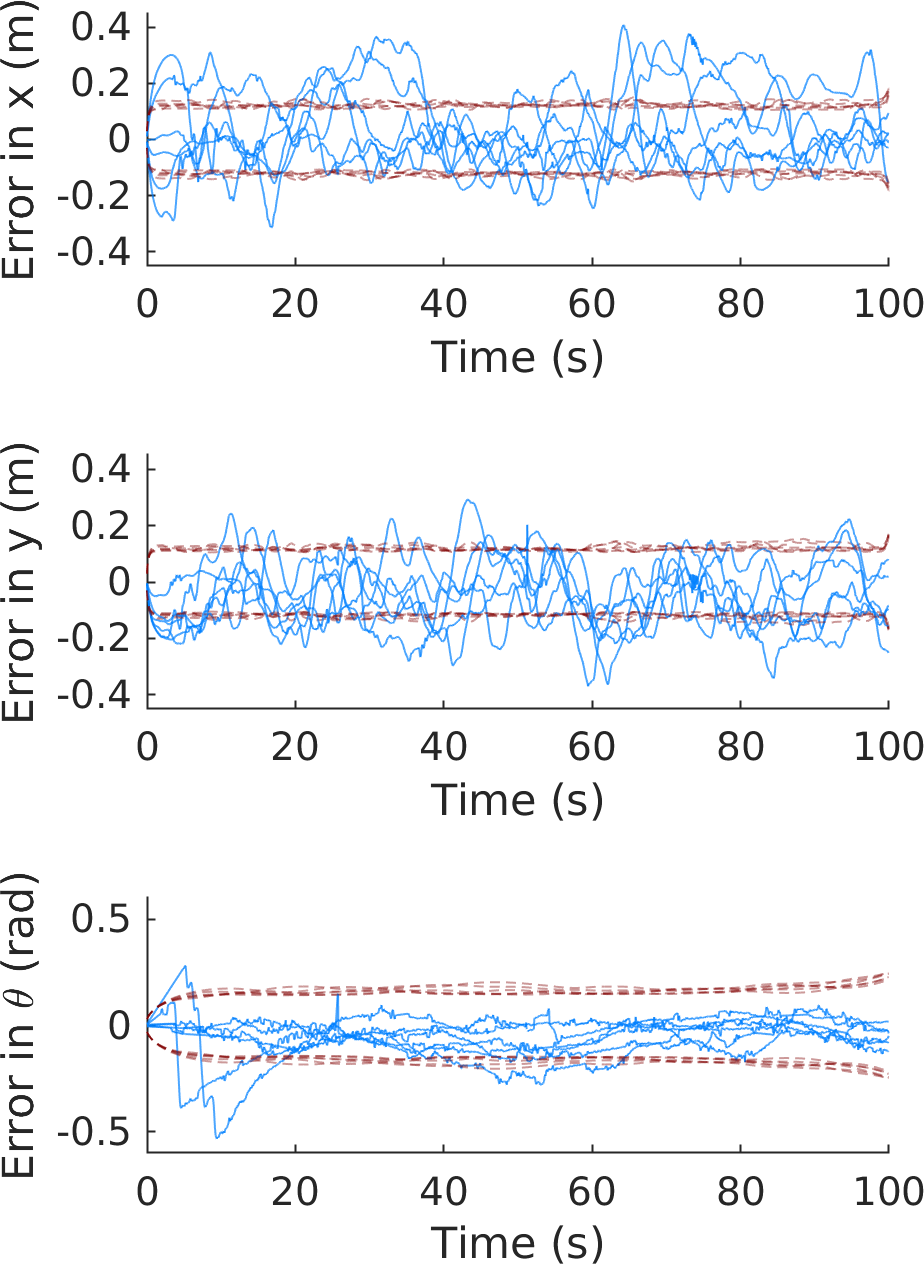}
 }
 \end{widepage}
 \caption{Error plots of the 6 folds of the UWB experimental dataset for KoopSE (left) and the model-based extended RTS smoother (right). The blue lines represent the errors of the estimated trajectories, and the red envelopes represent the estimated $3\sigma$ bounds. The errors of KoopSE are smaller and bounded by the $3\sigma$ bounds, while those of the model-based smoother are larger and often not bounded.\vspace{10pt}}
 \label{fig:dataset-errors}
\end{figure}
Experiments in a lab setting using a Clearpath Husky Unmanned Ground Vehicle (UGV) were used to validate the proposed approach. A 30-minute dataset of the UGV driving in an indoor environment was collected. Ground-truth position and orientation data was collected using an OptiTrack motion capture system. The wheel odometry consisting of forward velocity and yaw rate was calculated from wheel encoders. Five UWB anchors transmitting range measurements were used, with two of the anchors obstructed with metal plates representing clutter in the environment, creating additional measurement bias on the order of 30 cm. A picture of the Husky and the anchors is shown in \figref{fig:robot-pic}.

\begin{figure}[!t]
 \centering
%  \begin{widepage}
 \subcaptionbox*{}{
   \includegraphics[width=0.466\columnwidth]{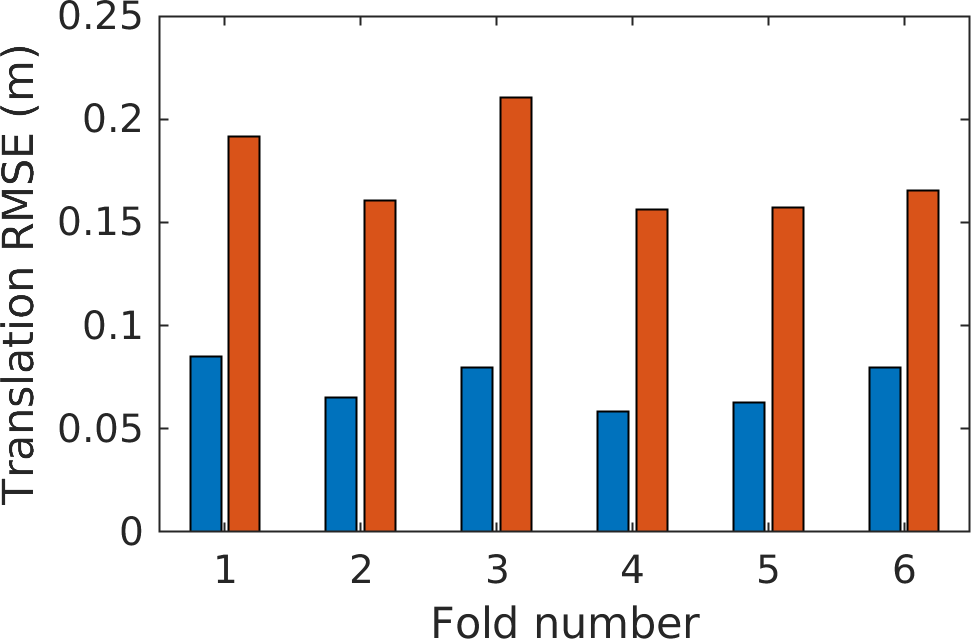}
   }
 \subcaptionbox*{}{
   \includegraphics[width=0.466\columnwidth]{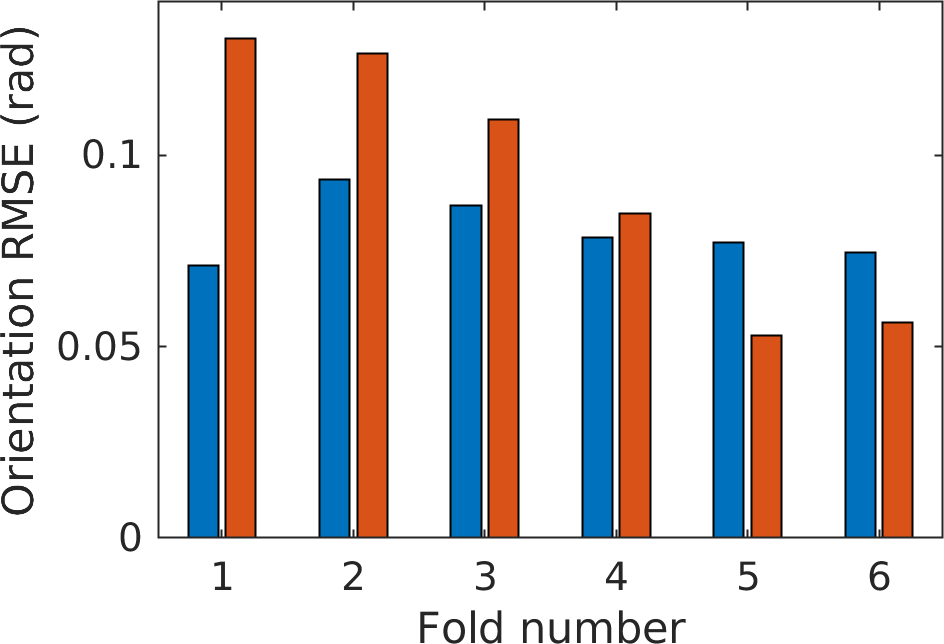}
}
 \subcaptionbox*{}{
   \includegraphics[width=0.466\columnwidth]{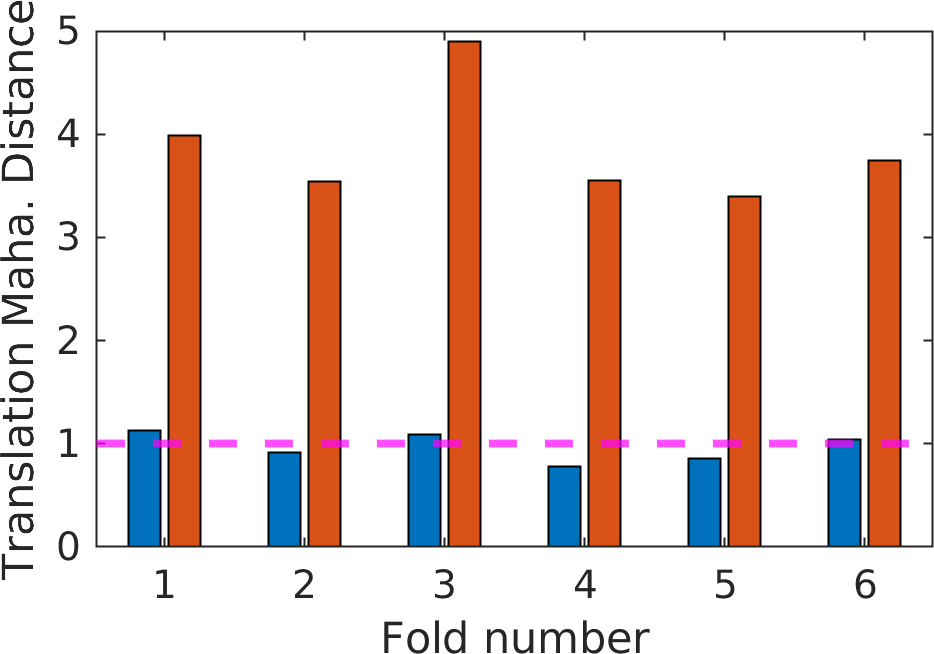}
   }
 \subcaptionbox*{}{
   \includegraphics[width=0.466\columnwidth]{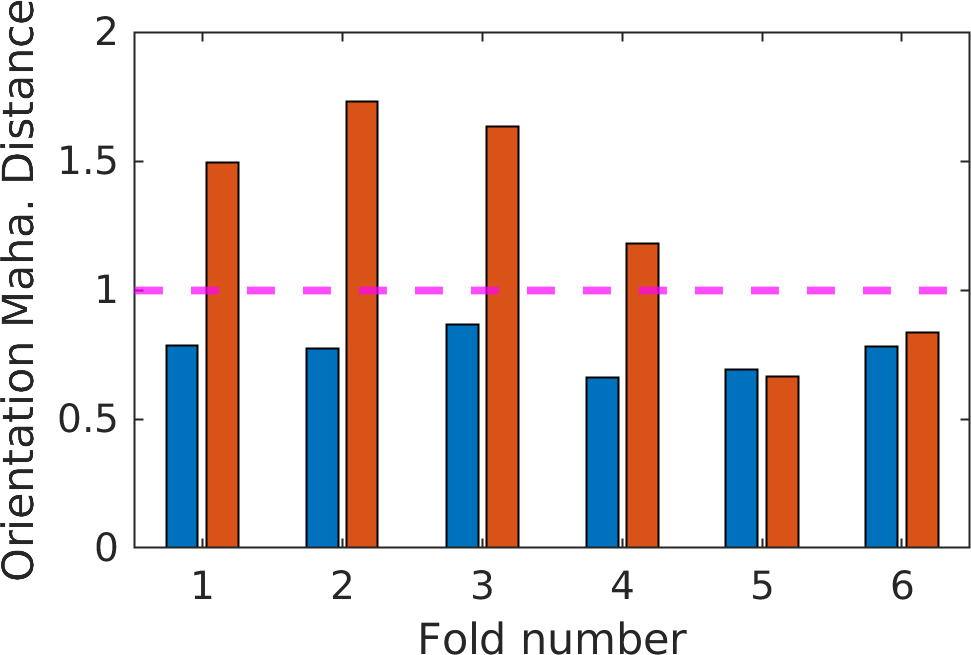}
}
% \end{widepage}
 \caption{RMSE and Mahanalobis distance for translation and orientation for KoopSE (blue) and the model-based extended RTS smoother (orange) on the 6 folds of the robot dataset. For translation errors, KoopSE has a lower RMSE and a more consistent Mahalanobis distance than the model-based smoother across all folds. For errors in orientation, which is less affected by the UWB range measurements, KoopSE performs just as well as the model-based smoother.}
 \label{fig:k-fold-errors}
\end{figure}

To validate the generality of KoopSE, we ran a 6-fold cross-validation, training on 25 minutes of data and testing on a random 100-second section of the remainder. The error plots for each fold are shown in \figref{fig:dataset-errors}. The RMSE and Mahalanobis distances for the folds are shown in \figref{fig:k-fold-errors}.

\section{Discussion and Conclusion}
\label{sec:discussion_conclusion}
The results highlight the benefits of the data-driven approach. \tabref{tab:simu-errors} and \figref{fig:simu-errors} show that in simulation, KoopSE has lower errors than the model-based extended RTS smoother, which is ignorant of the range biases, even though both methods are consistent. This shows that, despite having no knowledge of the robot's motion model, nor the UWB range measurement model, nor the position of the anchors, KoopSE outperforms the classical method when a small unmodelled measurement bias is introduced. In fact, when the bias is removed, we found that KoopSE performed just as well as the model-based smoother. In \figref{fig:simu-varying}, the RMSE of KoopSE quickly decreases over the first 100 RFF and the first 10000 training points. KoopSE surpassed the classical smoother with only 128 RFF and 10000 training points (about 8 minutes), making it feasible for real-world applications.

The results on the experimental dataset validated our findings. As shown in \figref{fig:k-fold-errors}, compared to the model-based smoother, KoopSE achieved lower translational RMSE and similar orientation RMSE for all folds. Unlike for the simulation setting, the covariance parameters for the model-based smoother achieving the lowest RMSE resulted in overconfident position estimates, despite best efforts in tuning. This is likely due to unmodelled effects in its motion or sensor models, which KoopSE overcame with its model-free approach. Overall, KoopSE is an efficient framework for data-driven state estimation of control-affine systems. We have demonstrated KoopSE's feasibility in a nonlinear UWB-tracking problem, and have shown that this system and potentially many others are approximately bilinear-Gaussian in a higher-dimensional space, permitting state estimation with familiar linear tools. This method is applicable for cases where the system models are unknown or have complicated noise distributions, such as in our scenario of indoor navigation with UWB sensors.

Although KoopSE requires no prior knowledge on system models, training requires ground-truth states as input. Future work could investigate ways of learning system models without exact ground-truth states. As well, since the learned lifted models have proven to be effective for state estimation, a natural extension is to use the same models for data-driven control under a similar framework.

% \input{sections/10_Acknowledgment}

% \input{sections/Temp_latex_for_video}

% \highlightReference{diff-filter}
\bibliographystyle{IEEEtran}
{
\singlespacing
\bibliography{refs}

\begin{thebibliography}{10}
\providecommand{\url}[1]{#1}
\csname url@samestyle\endcsname
\providecommand{\newblock}{\relax}
\providecommand{\bibinfo}[2]{#2}
\providecommand{\BIBentrySTDinterwordspacing}{\spaceskip=0pt\relax}
\providecommand{\BIBentryALTinterwordstretchfactor}{4}
\providecommand{\BIBentryALTinterwordspacing}{\spaceskip=\fontdimen2\font plus
\BIBentryALTinterwordstretchfactor\fontdimen3\font minus
  \fontdimen4\font\relax}
\providecommand{\BIBforeignlanguage}[2]{{%
\expandafter\ifx\csname l@#1\endcsname\relax
\typeout{** WARNING: IEEEtran.bst: No hyphenation pattern has been}%
\typeout{** loaded for the language `#1'. Using the pattern for}%
\typeout{** the default language instead.}%
\else
\language=\csname l@#1\endcsname
\fi
#2}}
\providecommand{\BIBdecl}{\relax}
\BIBdecl

\bibitem{diff-filter}
R.~Jonschkowski, D.~Rastogi, and O.~Brock, ``Differentiable particle filters:
  End-to-end learning with algorithmic priors,'' in \emph{Proceedings of RSS},
  June 2018.

\bibitem{control-affine-to-bilin}
D.~Bruder, X.~Fu, and R.~Vasudevan, ``Advantages of bilinear {Koopman}
  realizations for the modeling and control of systems with unknown dynamics,''
  \emph{IEEE RAL}, vol.~6, no.~3, pp. 4369--4376, 2021.

\bibitem{Hofmann}
T.~Hofmann, B.~Schölkopf, and A.~J. Smola, ``{Kernel methods in machine
  learning},'' \emph{Ann. Stat.}, vol.~36, no.~3, p. 1171–1220, Jun 2008.

\bibitem{song}
L.~Song, J.~Huang, A.~Smola, and K.~Fukumizu, ``Hilbert space embeddings of
  conditional distributions with applications to dynamical systems,'' in
  \emph{Proceedings of the 26th ICML}, ser. ICML '09.\hskip 1em plus 0.5em
  minus 0.4em\relax New York, NY, USA: Association for Computing Machinery,
  2009, p. 961–968.

\bibitem{kbr}
K.~Fukumizu, L.~Song, and A.~Gretton, ``{Kernel Bayes' rule: Bayesian inference
  with positive definite kernels},'' \emph{JMLR}, vol.~14, no.~82, pp.
  3753--3783, 2013.

\bibitem{ksmoother}
Y.~Nishiyama, A.~Afsharinejad, S.~Naruse, B.~Boots, and L.~Song, ``{The
  nonparametric kernel Bayes smoother},'' in \emph{AISTATS}, 2016.

\bibitem{kkr}
G.~Gebhardt, A.~Kupcsik, and G.~Neumann, ``{The kernel Kalman rule: efficient
  nonparametric inference by recursive least-squares and subspace
  projections},'' \emph{Machine Learning}, vol. 108, 06 2019.

\bibitem{gp-book}
C.~E. Rasmussen and C.~K.~I. Williams, \emph{Gaussian Processes for Machine
  Learning}.\hskip 1em plus 0.5em minus 0.4em\relax Cambridge, MA, USA: MIT
  Press, 2005.

\bibitem{GPBayes}
J.~{Ko} and D.~{Fox}, ``{GP-BayesFilters: Bayesian filtering using Gaussian
  process prediction and observation models},'' in \emph{2008 IEEE/RSJ IROS},
  2008, pp. 3471--3476.

\bibitem{GPnoiseMulti}
L.~{McCalman}, S.~{O'Callaghan}, and F.~{Ramos}, ``{Multi-modal estimation with
  kernel embeddings for learning motion models},'' in \emph{2013 IEEE ICRA},
  2013, pp. 2845--2852.

\bibitem{Koopman}
B.~O. Koopman, ``{Hamiltonian systems and transformation in Hilbert space},''
  \emph{PNAS}, vol.~17, no.~5, pp. 315--318, 1931.

\bibitem{mauroy_2020_koopman}
A.~Mauroy, I.~Mezi{\'{c}}, and Y.~Susuki, \emph{The {Koopman} Operator in
  Systems and Control}.\hskip 1em plus 0.5em minus 0.4em\relax {New York, NY,
  USA}: Springer Publishing, 2020.

\bibitem{dmd-book}
J.~Kutz, S.~Brunton, B.~Brunton, and J.~Proctor, \emph{{Dynamic mode
  decomposition: data-driven modeling of complex systems}}.\hskip 1em plus
  0.5em minus 0.4em\relax {Philadelphia, PA, USA}: {SIAM}, 2016.

\bibitem{dmd-big-book}
S.~L. Brunton and J.~N. Kutz, \emph{Data-Driven Science and Engineering:
  Machine Learning, Dynamical Systems, and Control}.\hskip 1em plus 0.5em minus
  0.4em\relax Cambridge, U.K.: Cambridge Univ. Press, 2019.

\bibitem{koopman-so3}
T.~Chen and J.~Shan, ``{Koopman-operator-based attitude dynamics and control on
  SO(3)},'' \emph{J. Guid. Control Dyn.}, vol.~43, 09 2020.

\bibitem{koopman-control}
I.~Abraham and T.~Murphey, ``{Active learning of dynamics for data-driven
  control using Koopman operators},'' \emph{IEEE Trans. Robot.}, vol.~35, pp.
  1071--1083, 2019.

\bibitem{rff}
A.~Rahimi and B.~Recht, ``Random features for large-scale kernel machines,'' in
  \emph{NIPS}, J.~Platt, D.~Koller, Y.~Singer, and S.~Roweis, Eds.,
  vol.~20.\hskip 1em plus 0.5em minus 0.4em\relax Curran Associates, Inc.,
  2008.

\bibitem{hilbert-maps}
F.~Ramos and L.~Ott, ``Hilbert maps: scalable continuous occupancy mapping with
  stochastic gradient descent,'' \emph{IJRR}, vol.~35, no.~14, pp. 1717--1730,
  2016.

\bibitem{rff-learn-dynamics}
A.~Gijsberts and G.~Metta, ``Incremental learning of robot dynamics using
  random features,'' \emph{IEEE ICRA}, pp. 951--956, 05 2011.

\bibitem{koopman-with-rff}
A.~M. DeGennaro and N.~M. Urban, ``Scalable extended dynamic mode decomposition
  using random kernel approximation,'' \emph{SIAM Journal on Scientific
  Computing}, vol.~41, no.~3, pp. A1482--A1499, 2019.

\bibitem{rkhs}
A.~Smola, A.~Gretton, L.~Song, and B.~Sch{\"o}lkopf, ``A hilbert space
  embedding for distributions,'' in \emph{Algorithmic Learning Theory},
  M.~Hutter, R.~A. Servedio, and E.~Takimoto, Eds.\hskip 1em plus 0.5em minus
  0.4em\relax Berlin, Heidelberg: Springer Berlin Heidelberg, 2007, pp. 13--31.

\bibitem{rkhs-book}
J.~H. Manton and P.-O. Amblard, \emph{A Primer on Reproducing Kernel Hilbert
  Spaces}.\hskip 1em plus 0.5em minus 0.4em\relax Norwell, MA, USA: Now
  Publishers, 2015, vol.~8, no. 1–2.

\bibitem{kernel-embeddings}
K.~Muandet, K.~Fukumizu, B.~Sriperumbudur, and B.~Schölkopf, \emph{Kernel Mean
  Embedding of Distributions: A Review and Beyond}.\hskip 1em plus 0.5em minus
  0.4em\relax Norwell, MA, USA: Now Publishers, 2017, vol.~10.

\bibitem{representer-thm}
B.~Sch{\"o}lkopf, R.~Herbrich, and A.~J. Smola, ``A generalized representer
  theorem,'' in \emph{Computational Learning Theory}, D.~Helmbold and
  B.~Williamson, Eds.\hskip 1em plus 0.5em minus 0.4em\relax Berlin,
  Heidelberg: Springer, 2001, pp. 416--426.

\bibitem{esgvi-extended}
J.~N. Wong, D.~J. Yoon, A.~P. Schoellig, and T.~D. Barfoot, ``Variational
  inference with parameter learning applied to vehicle trajectory estimation,''
  \emph{IEEE RA-L}, vol. abs/2003.09736, 2020.

\bibitem{barfoot-txtbk}
T.~D. Barfoot, \emph{State Estimation for Robotics}.\hskip 1em plus 0.5em minus
  0.4em\relax {Cambridge, U.K.}: {Cambridge Univ. Press}, 2017.

\bibitem{probabilistic-robotics}
S.~Thrun, W.~Burgard, and D.~Fox, \emph{Probabilistic Robotics}.\hskip 1em plus
  0.5em minus 0.4em\relax Cambridge, MA, USA: {The MIT Press}, 2005.

\bibitem{rff-periodic}
A.~Tompkins and F.~Ramos, ``Fourier feature approximations for periodic kernels
  in time-series modelling,'' in \emph{AAAI}, 2018.

\bibitem{wang-13}
F.~Wang and A.~E. Gelfand, ``Directional data analysis under the general
  projected normal distribution,'' \emph{Statistical Methodology}, vol.~10,
  no.~1, pp. 113--127, 2013.

\end{thebibliography}
}
\appendix
\subsection{Detailed Sketch that KoopSE is Kernelized}
\label{sec:kernel-proof-appendix}
Our goal is to show that KoopSE uses the RKHS quantities only in their kernelized forms.  For training, the lifted training points are used to compute the bilinear system matrices in \eqref{eq:bilin-stochastic}. Using SMW identities, we can write the analytical solution of \eqref{eq:abhcqr} for $\mbf{A}$, $\mbf{B}$, and $\mbf{H}$, as well as an alternative expression for $\mbf{C}$:
\begin{subequations}
\begin{alignat}{2}
 \mbf{A} &= \frac{1}{\lambda_A} \mbf{X} \mbf{L} \tilde{\mbf{X}}^T, \quad
 &\mbf{B} &= \frac{1}{\lambda_B} \mbf{X} \mbf{L} \mbf{U}^T, \\
 \mbf{H} &= \frac{1}{\lambda_H} \mbf{X} \mbf{L} \mbf{V}^T, \quad
 &\mbf{C} &= \mbf{Y} \left( \mbf{X}^T \mbf{X} + \lambda_C \mbf{1} \right)^{-1} \mbf{X}^T,
\end{alignat}
\end{subequations}
where $\mbf{L} = \left( \frac{1}{\lambda_A} \tilde{\mbf{X}}^T \tilde{\mbf{X}} + \frac{1}{\lambda_B} \mbf{U}^T \mbf{U} + \frac{1}{\lambda_H} \mbf{V}^T \mbf{V} + \mbf{1}\right)^{-1}$. This solution is not used in practice is because it involves inverting $P\times P$ matrices, where $P$ is the amount of training data. However, the solution is clearly of the form
\begin{subequations}
\begin{alignat}{2}
\mbf{A} & = \mbf{X} \mbf{W}_A \mbf{X}^T, \quad
&\mbf{B} &= \mbf{X} \mbf{W}_B \mbf{U}^T, \\
\mbf{H} & = \mbf{X} \mbf{W}_H \mbf{V}^T, \quad
&\mbf{C} &= \mbf{Y} \mbf{W}_C \mbf{X}^T, \\
 \mbf{Q} & = \mbf{X} \mbf{W}_Q \mbf{X}^T + \lambda_Q \mbf{1}, \quad
&\mbf{R} &= \mbf{Y} \mbf{W}_R \mbf{Y}^T + \lambda_R \mbf{1},
\end{alignat}
\end{subequations}
where $\mbf{W}_A, \mbf{W}_B, \mbf{W}_H, \mbf{W}_C, \mbf{W}_Q, \mbf{W}_R$ are some kernelized matrices.

At test time, we assume that the initial condition, new control inputs, and new measurements can be written as linear combinations of those seen in training:
\begin{subequations}
 \begin{align}
  \check{\mbf{x}}^\prime_0 &= \mbf{X}\mbf{w}_{\check{x},0}, \\
  {\mbf{u}}^\prime_k &= \mbf{U}\mbf{w}_{u,k}, \quad k = 1,\dots,K \\
  {\mbf{y}}^\prime_k &= \mbf{Y}\mbf{w}_{y,k}, \quad k = 0,\dots,K
 \end{align}
 \end{subequations}
for weights $\mbf{w}_{\check{x},0}$, $\mbf{w}_{u,k}$, and $\mbf{w}_{y,k}$. Looking at the definition of $\mbf{A}_{k-1}$ in \eqref{eq:convert-to-ltv}, we notice that
\begin{subequations}
\label{eq:h-kernelized}
{\small
\begin{align}
 &\mbf{H}(\mbf{u}_k^\prime \otimes \mbf{1}) = \mbf{X}\mbf{W}_H(\mbf{U}\odot\mbf{X})^T(\mbf{u}_k^\prime \otimes \mbf{1}) \\
 &= \mbf{X}\mbf{W}_H
 \begin{bmatrix}
  (\mbf{u}^{(1)T}\otimes\mbf{x}^{(1)T})(\mbf{u}_k^\prime\otimes\mbf{1}) \\
  \vdots \\
  (\mbf{u}^{(P)T}\otimes\mbf{x}^{(P)T})(\mbf{u}_k^\prime\otimes\mbf{1})
  \end{bmatrix} \\
 &= \mbf{X}\mbf{W}_H
 \begin{bmatrix}
  ((\mbf{u}^{(1)T}\mbf{u}_k^\prime)\otimes\mbf{1})\mbf{x}^{(1)T} \\
  \vdots \\
  ((\mbf{u}^{(P)T}\mbf{u}_k^\prime)\otimes\mbf{1})\mbf{x}^{(P)T} 
 \end{bmatrix} \\
 &= \mbf{X}\mbf{W}_H
 \begin{bmatrix}
  (\mbf{u}^{(1)T}\mbf{u}_k^\prime)\otimes\mbf{1} & & \\
  & \ddots & \\
  & & (\mbf{u}^{(P)T}\mbf{u}_k^\prime)\otimes\mbf{1}
 \end{bmatrix} \mbf{X}^T
 \label{eq:h-kernelized-block-matrix} \\
 &= \mbf{X} \mbf{W}_{H,k} \mbf{X}^T
\end{align}
}
\end{subequations}
\\[-10pt]  % weird space after \small if line break is not used
where $\mbf{W}_{H,k}$ is the product of $\mbf{W}_H$ and the kernelized block diagonal matrix in \eqref{eq:h-kernelized-block-matrix}. Then, for $k=0,\dots,K$,
\begin{subequations}
\begin{align}
 \mbf{A}_{k-1} &= \mbf{A} + \mbf{H}(\mbf{u}_k^\prime \otimes \mbf{1}) \\
 &= \mbf{X}\mbf{W}_{A}\mbf{X}^T + \mbf{X}\mbf{W}_{H,k}\mbf{X}^T \\
 &= \mbf{X}\mbf{W}_{A,k}\mbf{X}^T,
 \end{align}
 \end{subequations}
 \begin{subequations}
 \begin{align}
 \mbf{v}_k^\prime &= \mbf{B}\mbf{u}_k^\prime \\
 &= (\mbf{X}\mbf{W}_B\mbf{U}^T)(\mbf{U}\mbf{w}_{u,k}) \\
 &= \mbf{X}\mbf{w}_{v,k},
\end{align}
\end{subequations}
 where
 \begin{gather}
  \mbf{W}_{A,k} = \mbf{W}_A + \mbf{W}_{H,k}, \quad
  \mbf{w}_{v,k} = \mbf{W}_B(\mbf{U}^T\mbf{U})\mbf{w}_{u,k}
 \end{gather}
are clearly kernelized. Now, the solution to \eqref{eq:ltv-system} is exactly solved by the RTS smoother \cite{barfoot-txtbk}. We can then look at the structure of the estimated state means and covariances of the forward pass, then of backward pass, through induction. We outline the setup below. Let $\{\check{\mbf{x}}_{k,f},\check{\mbf{P}}_{k,f}\}_{k=0}^K$, $\{\hat{\mbf{x}}_{k,f},\hat{\mbf{P}}_{k,f}\}_{k=0}^K$, and $\{\hat{\mbf{x}}_{k},\hat{\mbf{P}}_{k}\}_{k=0}^K$ denote the means and covariances of the forward pass prior (i.e., dead reckoning), forward pass posterior (i.e., Kalman Filter), and the backward pass (i.e., smoother solutions), respectively. For the forward pass, the initial condition is
\begin{gather}
 \check{\mbf{x}}_0^\prime = \mbf{X}\mbf{w}_{\check{x},0}, \quad
 \check{\mbf{P}}^\prime_0 = \mbf{Q} = \mbf{X}\mbf{W}_Q\mbf{X}^T+\lambda_Q\mbf{1}.
\end{gather}
Suppose at timestep $k$, the priors have the form
\begin{gather}
\label{eq:fpass-structure}
 \check{\mbf{x}}_k^\prime = \mbf{X}\mbf{w}_{\check{x},k,f}, \quad
 \check{\mbf{P}}^\prime_k = \mbf{X}\mbf{W}_{\check{P},k,f}\mbf{X}^T + \check{c}_{k,f}\mbf{1},
\end{gather}
for some kernelized matrices $\mbf{w}_{\check{x},k,f}$ and $\mbf{W}_{\check{P},k,f}\mbf{X}^T$ and scalar $\check{c}_{k,f}$. Then, going through the forward pass and using the derived kernelized structures for $\mbf{A}_{k-1}$, $\mbf{C}$, $\mbf{Q}$, $\mbf{R}$, $\mbf{v}_k^\prime$, and $\mbf{y}_k^\prime$, it is straightforward to show that
\begin{gather}
  \hat{\mbf{x}}_k^\prime = \mbf{X}\mbf{w}_{\hat{x},k,f}, \quad
 \hat{\mbf{P}}^\prime_k = \mbf{X}\mbf{W}_{\hat{P},k,f}\mbf{X}^T + \check{c}_{k,f}\mbf{1},
\end{gather}
and then that
\begin{gather}
  \check{\mbf{x}}_{k+1}^\prime = \mbf{X}\mbf{w}_{\check{x},k+1,f}, \quad
 \check{\mbf{P}}^\prime_{k+1} = \mbf{X}\mbf{W}_{\check{P},k+1,f}\mbf{X}^T + \check{c}_{k+1,f}\mbf{1}
\end{gather}
with the same matrix structures. The proof setup is similar for the backward pass. Through these induction processes, we can show that the mean and covariance outputs from the forward pass prior, forward pass posterior, and the backward pass all have the structure, for $k=0,\dots,K$,
\begin{align}
 \hat{\mbf{x}}_k^\prime = \mbf{X}\mbf{w}_{\hat{x},k}, \quad
 \hat{\mbf{P}}_k^\prime = \mbf{X}\mbf{W}_{\hat{P},k}\mbf{X}^T + c_k\mbf{1},
\end{align}
with kernelized matrices $\mbf{w}_{\hat{x},k}$ and $\mbf{W}_{\hat{P},k}$ and scalar $c_k$. We substitute these expressions for $\hat{\mbf{x}}_k^\prime$ and $\hat{\mbf{P}}_k^\prime$ into \eqref{eq:outputs}, yielding
\begin{subequations}
\begin{align}
 \hat{\mbs{\xi}}_k^\prime &= \mbs{\Xi} (\mbf{X}^T \mbf{X} + \lambda_x \mbf{1})^{-1} \mbf{X}^T (\mbf{X}\mbf{w}_{x,k}), \\
 \hat{\mbs{\Sigma}}_k^\prime &= \mbs{\Xi} (\mbf{X}^T \mbf{X} + \lambda_x \mbf{1})^{-1} \mbf{X}^T (\mbf{X}\mbf{W}_{P,k}\mbf{X}^T+{c}_k\mbf{1}) \nonumber \\
 &\times\mbf{X}(\mbf{X}^T \mbf{X} + \lambda_x \mbf{1})^{-1} \mbs{\Xi}^T.
\end{align}
\end{subequations}
We can see that the results for $\hat{\mbs{\xi}}_k^\prime$ and $\hat{\mbs{\Sigma}}_k$, the states and covariances in the original space, are indeed kernelized. Therefore, from input to output, the algorithm only uses RKHS quantities in their kernelized forms.

\end{document}